\documentclass[AMA,STIX1COL]{WileyNJD-v2}
\usepackage{moreverb}
\usepackage{algorithm}
\usepackage{algpseudocode}
\usepackage{todonotes}

\algblock{Input}{EndInput}
\algnotext{EndInput}
\algblock{Output}{EndOutput}
\algnotext{EndOutput}

\newcommand\BibTeX{{\rmfamily B\kern-.05em \textsc{i\kern-.025em b}\kern-.08em
T\kern-.1667em\lower.7ex\hbox{E}\kern-.125emX}}

\articletype{Original Article (Under review)}%

\received{<day> <Month>, <year>}
\revised{<day> <Month>, <year>}
\accepted{<day> <Month>, <year>}

\begin{document}

\title{Automatic tuning of hyper-parameters of reinforcement learning algorithms using Bayesian optimization with behavioral cloning\protect}

\author[1]{Juan Cruz Barsce}

\author[1,2,3]{Jorge A. Palombarini}

\author[4]{Ernesto C. Martínez}

\authormark{BARSCE \textsc{et al}}

\address[1]{\orgdiv{Dpt. of Information Systems Engineering, Facultad Regional Villa María}, \orgname{Universidad Tecnológica Nacional (UTN)}, \orgaddress{\state{Villa María}, \country{Argentina}}}

\address[2]{\orgdiv{GISIQ, Facultad Regional Villa María}, \orgname{UTN}, \orgaddress{\state{Villa María}, \country{Argentina}}}

\address[3]{\orgdiv{CIT Villa María}, \orgname{CONICET-UNVM}, \orgaddress{\state{Villa María}, \country{Argentina}}}

\address[4]{\orgdiv{INGAR}, \orgname{CONICET-UTN}, \orgaddress{\state{Santa Fe}, \country{Argentina}}}

\corres{Ernesto C. Martínez, Avellaneda 3657, INGAR, S3002GJC, Santa Fe, Argentina. \email{ecmarti@santafe-conicet.gob.ar}}


\abstract[Abstract]{Optimal setting of several hyper-parameters in machine learning algorithms is key to make the most of available data. To this aim, several methods such as evolutionary strategies, random search, Bayesian optimization and heuristic rules of thumb have been proposed. In reinforcement learning (RL), the information content of data gathered by the learning agent while interacting with its environment is heavily dependent on the setting of many hyper-parameters. Therefore, the user of an RL algorithm has to rely on search-based optimization methods,  such as grid search or the Nelder-Mead simplex algorithm, that are very inefficient for most RL tasks, slows down significantly the learning curve and leaves to the user the burden of purposefully biasing data gathering. In this work, in order to make an RL algorithm more user-independent, a novel approach for autonomous hyper-parameter setting using Bayesian optimization is proposed. Data from past episodes and different hyper-parameter values are used at a meta-learning level by performing behavioral cloning which helps improving the effectiveness in maximizing a reinforcement learning variant of an acquisition function. Also, by tightly integrating Bayesian optimization in a reinforcement learning agent design, the number of state transitions needed to converge to the optimal policy for a given task is reduced. Computational experiments reveal promising results compared to other manual tweaking and optimization-based approaches which highlights the benefits of changing the algorithm hyper-parameters to increase the information content of generated data.}

\keywords{Reinforcement learning; Hyper-parameter optimization; Bayesian optimization; Behavioral cloning}

\maketitle

\section{Introduction} 
\label{sec:introduction}

The process of induction in supervised learning algorithms involves generalizing from training data in order to provide the best prediction for unseen data during testing. Typically, a learning algorithm usually has parameters that are fitted during training (e.g., weight parameters in a neural network) and hyper-parameters (e.g., the structure of a neural network or the learning rate $\alpha_{lr}$) that make assumptions about the data and impose bias before beginning the training phase. Such bias severely constraints what can be learned and the quality of the resulting predictive model. As the training data set does not change during training, hyper-parameter optimization can be carried out independently from learning. In order to ensure effective learning of an inductive model and, therefore, to provide a better generalization to new data, the task of setting the learning algorithm hyper-parameters is crucial, being responsible for either a mediocre or an outstanding generalization capability of the fitted model \cite{hutter_beyond_2015}. This has motivated a considerable research effort to address hyper-parameter optimization using different approaches such as Tree-structured Parzen Estimators (TPE) \cite{bergstra_algorithms_2011}, grid and random search \cite{bergstra_random_2012} Bayesian optimization \cite{snoek_practical_2012}, reinforcement learning for neural network architecture search \cite{zoph_neural_2016}, Covariance Matrix Adaptation Evolution Strategy (CMA-ES) \cite{loshchilov_cma-es_2016}, to name but a few.

As an optimal control problem under uncertainty, in reinforcement learning (RL) \cite{sutton_reinforcement_2018}, the data set used by a learning agent to estimate an optimal policy is not provided \textit{a priori}. Instead, it is the direct result of a sequence of interactions between such an agent and its environment, from which it receives rewards associated with state transitions caused by the actions taken. The information content in the data generated by such interactions is biased by a number of algorithm hyper-parameters used during training. Hence, poor setting of these hyper-parameters negatively influences the learning curve and the gap between the policy learned and the optimal one.

The actions tried at visited environmental states aim to maximize the cumulative reward the agent will receive using the learned policy after training. To this aim, during training, the setting of hyper-parameters must carefully balance the trade-off between exploring what would be highly informative for approximating an optimal policy and exploiting what the agent already knows. Optimization of hyper-parameters is thus a kind of active meta-learning to influence the data set on which the optimal policy is going to be estimated.

There have been impressive successes in real-world applications of reinforcement learning, including agents that play different Atari games with human performance \cite{mnih_human-level_2015}, agents that excelled at the game of Go beating some of the best human players \cite{silver_mastering_2016,silver_mastering_2017-1}, and can play grand-master StarCraft 2 \cite{vinyals_grandmaster_2019}, robotics control \cite{lillicrap_continuous_2015}, among many others. However, regarding the setting of hyper-parameters used to accomplish these tasks using reinforcement learning, previous works either do not provide a detailed account on how the hyper-parameters were tuned and which alternative settings were considered in such implementations \cite{henderson_deep_2017}, or simply mentioned that the hyper-parameter configuration was obtained empirically or by \textit{"informal search"} (e.g., the latter is the case in Mnih et al. (2015) \cite{mnih_human-level_2015}). As stated above, lacking a systematic approach to optimize the hyper-parameters in the very design of RL algorithms is a serious drawback since the training data set is the direct result of the values chosen for the hyper-parameters, mainly those that bias which states are visited and those that generate the most informative \textit{(state, action)} pairs that define an optimal policy. Since there always exists a stochastic effect of actions taken at a given state, different configurations of hyper-parameters can yield considerable variability in the estimation of the optimal policy, as stated in Henderson et al. (2017) \cite{henderson_deep_2017} and Islam et al. (2017) \cite{islam_reproducibility_2017}.

In comparison with supervised learning, the experimental cost and risk of data gathering for reinforcement learning are factors that make exhaustive methods of hyper-parameter optimization very expensive and often inefficient. Thus, experimental design for actively seeking highly informative data often relies on informal tuning or randomized search. The main drawback of informal tuning is that it is often suboptimal, so that the performance of the configured RL algorithm is very inefficient, as it biases sampling state-action pairs without any concern for their information content. A poor setting of hyper-parameters may even lead to divergence in the search for an optimal policy, whereas the problem with randomized search is that it makes little use of the information gathered in the previous learning episodes. As a result, hyper-parameters in reinforcement learning are often heuristically tweaked, and there is no information about how the chosen values of hyper-parameters for algorithms used in different works affect the results obtained \cite{islam_reproducibility_2017}. Another issue with manual hyper-parameter tuning in reinforcement learning is that, considering that the true values of the states or alternative policies are a priori unknown, a learning agent often converges to a suboptimal policy.

Bearing in mind that many existing approaches mainly focus on supervised learning, this work specifically deals with active meta-learning using Bayesian optimization of the chosen values for the different hyper-parameters in a reinforcement learning algorithm. To this aim, an adaptation of Bayesian optimization with an extension of a common acquisition function, specifically geared for RL, is proposed. A novel two-step expected improvement function, particularly suited to autonomously optimize the hyper-parameters used by a reinforcement learning agent while solving a task by interacting with its environment is presented.

To enhance meta-learning of the algorithm hyper-parameters, behavioral cloning (BC) is used. BC is a method to learn policies by learning from a (typically small) set of demonstrations, casted as the supervised learning problem of minimizing the discrepancy between the selected action and the demonstrated action. Used in reinforcement learning as a mechanism to pre-train policies prior to starting using an RL algorithm, BC reduces the samples required to converge towards the optimal policy \cite{hester_deep_2017}. In this work, it is shown  that the integration of BC increases the efficiency of hyper-parameter setting, which makes room for resorting to the experience gathered with different hyper-parameters as demonstrations.

This work is organized as follows: Section \ref{sec:background} presents the methodologies used to define the meta-learning layer, Section \ref{sub:related_work_in_hyper_parameter_optimization} presents the related works, and Section \ref{sec:proposed_method} presents the proposed method. On the other hand, Section \ref{sec:experimental} validates and discusses a number of examples where the proposed method outperforms other state-of-art methods for a task. Meta-learning addresses the optimization of hyper-parameters of a reinforcement learning agent in five PyBullet  simulated robotic tasks \cite{coumans2019} and in the BSuite \cite{osband_behaviour_2019} collection of environments. Finally, the conclusions and future works are stated in Section \ref{sec:conclusion_and_future_works}.


\section{Methodology} 
\label{sec:background}

\subsection{Reinforcement Learning}

Reinforcement learning (RL) \cite{sutton_reinforcement_2018} is an area of machine learning that has put forward the possibilities of computational intelligence in domains such as games \cite{mnih_human-level_2015,OpenAI_dota}, self-driving cars \cite{shalev-shwartz_safe_2016} and dialogue generation \cite{li_deep_2016}. RL deals with an autonomous agent situated in an environment that attempts to learn a policy that maximizes the cumulative reward it may get, by performing a sequence of actions and observing the rewards associated with state transitions resulting from actions taken. To this aim, an evaluative feedback mechanism or learning rule is used. The problem is formalized as a Markov Decision Process (MDP) $(S, A, R, P, \gamma)$ \cite{sutton_reinforcement_2018} where:

\begin{itemize}
	\item $S$, is a set of states.
	\item $A$, is a set of actions.
	\item $R$, is a reward function that determines the rewards associated with state transitions due to actions taken.
	\item $P(s_{t+1}=s' \mid s,a)$, is a probability for the agent to transition to a certain state $s'$ at the next time-step, being in a state $s$ at time-step $t$, when an action $a$ is taken.
	\item $\gamma \in [0, 1)$, is a scalar that imposes a discount for future rewards, determining how short-- or far -- sighted is the policy of the agent. At time-step $t$, the cumulative discounted return is given by
	\begin{equation}
		\label{eqn_cumulative_rewards}
		R_t = r_t + \gamma r_{t+1} + \gamma^2 r_{t+2} + \dots
	\end{equation}
	It is worth noting that $\gamma$ is constrained to values less than $1$, as a necessary condition for the convergence of the learning algorithm.
\end{itemize}

The policy is defined as a function $\pi(a \mid s)$, and represents the probability of taking an action $a$, being in a state $s$. With $\pi$, the agent aims to maximize the \textit{value function} for every state; such function is defined as the expected reward, starting from a given state at time-step $t$, and following policy $\pi$ thereafter. Formally, this is given by the Bellman Equation:

\begin{eqnarray}
	\label{eqn:bellman_eq}
	V_\pi(s) &=& \mathbb{E}(R_t \mid s_t = s) \nonumber\\
	&=& \mathbb{E}(r_t + \gamma r_{t+1} + \gamma^2 r_{t+2} + \dots \mid s_t = s) \nonumber\\
	&=& \mathbb{E}(r_t + \gamma V_{\pi}(s_{t+1}) \mid s_t = s)
\end{eqnarray}

Analogously, the $Q$-value function extends Eqn. \ref{eqn:bellman_eq} to state-action pairs, and it is given by

\begin{eqnarray}
	\label{eqn:bellman_eq_q}
	Q(s,a) = \mathbb{E}(R_t \mid s_t = s, a_t = a)
\end{eqnarray}

The optimal policy is defined as a policy $\pi^*$ that is as good as or better than any other policy $\pi$, i.e., $V_{\pi^*}(s) \geq V_{\pi}(s), \forall s$.

Using RL to find an optimal policy given an MDP involves algorithms that are focused on either improving the value function $V(s)$, such as temporal difference learning, or algorithms that focus directly on learning the policy $\pi(a \mid s)$ such as REINFORCE, among others (see Sutton and Barto (2018) \cite{sutton_reinforcement_2018} for more information).

A basic algorithm in RL for solving MDPs is $Q$-Learning \cite{watkins_q-learning_1992}, which takes actions suggested by any policy (not necessarily the optimal one) and resorts to state transition and observed reward to update the value $Q(s,a)$ of a state-action pair using the rule:

\begin{equation}
	\label{eqn:q-learning}
	Q(s,a) \gets Q(s,a) + \alpha_{lr} (r + \gamma \arg\max_{a'} Q(s',a') - Q(s,a))
\end{equation}

As the action from which the next state value is discounted in the update is the best estimated action based on current knowledge, which is not determined by the policy being followed, it is said that $Q$-learning is an \textit{off-policy} algorithm. If the action would be determined by the policy being followed, then the algorithm is called \textit{on-policy}.

Key to properly addressing the reinforcement learning problem is the trade-off between exploration and exploitation. This gives rise to a dilemma in which the agent has to choose between taking actions that are considered to be the best according to the current estimation of the optimal policy, or taking actions that are deemed as suboptimal but makes room for the agent to discover better actions to exploit in the future. A simple approach to balance exploration with exploitation is the $\epsilon$-greedy policy, which determines that, at any state, the agent will take a random action (explore) with a given probability $\epsilon$, and execute the action regarded as best (exploit) with probability $(1-\epsilon)$.

When the state space is large, it is not efficient or even possible to update the value function via Equation \ref{eqn:q-learning}. Instead, the value function is approximated by a function parameterized with weights $\phi$, i.e., $V_\pi(s) \approx \hat{V}_\pi(s \mid \phi)$, commonly from a neural network model.

A notable extension of Q-learning is the Deep Q Learning algorithm (DQN) \cite{mnih_human-level_2015}, that extends its application scope using deep neural networks. This extension consists mainly in:

\begin{itemize}
	\item Predictions with action-value function $\hat{Q}(s,a \mid \phi)$ takes advantage of a deep neural network that can have many convolutional layers if the input state is an image.
	\item To prevent both prediction and target $y$ (defined below) to change at the same time in each gradient update, the target uses weights $\phi^-$ that are not updated by gradient descent, but replaced by weights $\phi$ every $C$ steps.
	\item A replay memory is used to store tuples of experience $(s, a, r, s')$ which are used for an off-policy estimation of the target $y$, given by $r$ if $s'$ is an absorbing state; otherwise, it is given by $r + \gamma max_{a'} \hat{Q} (s', a' \mid \phi^-)$. In the initial proposal of DQN, such tuples were uniformly sampled. This was further improved in Schaul et al. (2015) \cite{schaul_prioritized_2015-1}, when transitions are sampled with a probability related to the magnitude of the temporal-difference error.
	Then, the probability of selecting a transition $j$ is given by
	
	\begin{equation}
		P(j) = p_j^\alpha / \sum_i p_i^\alpha
	\end{equation}
	where the hyper-parameter $\alpha \in (0,1)$ sets a balance between sampling the transition with highest priority versus sampling a transition at random. Sampling in this way introduces a bias, that makes the expectation distribution  different from the distribution from which past experiences were sampled. To address this, updates on the network are corrected by using importance sampling weights given by
	
	\begin{equation}
		w_i = (N P(j))^{-\beta}
	\end{equation}
	where $N$ is the size of the experience replay buffer, and $\beta$ is a hyper-parameter that controls how the importance sampling correction affects value learning. More details regarding prioritizing experience replay can be found in Schaul et al. (2015) \cite{schaul_prioritized_2015-1}.
	
\end{itemize}

Another main family of RL algorithms is the \textit{policy gradient algorithms}. Instead of selecting the policy via $\epsilon-greedy$ given the estimation of $Q$ as in Q-Learning or DQN, they directly update the policy parameterized with weights $\theta$, i.e., $\pi(s,a \mid \theta)$, in order to maximize the received returns from following such policy (therefore, such algorithms are on-policy). Formally, given $\pi_\theta$, a stochastic policy parameterized with $\theta$, the objective is the maximization of the expected return,

\begin{equation}
	\label{eqn_expected_return}
	J(\pi_\theta) = \mathbb{E}_{\tau \sim \pi_\theta} [R(\tau)]
\end{equation}
where $R(\tau)$ is the sum of discounted rewards gathered in the sampled trajectory $\tau = (s_0, a_0, \dots, s_{t+1})$. The parameters $\theta$ of the policy $\pi_\theta$ can be updated via gradient ascent, such that the next update of the policy weights, $\theta_{t+1}$, is changed   in the improvement direction of the agent policy, controlled by $\alpha_{lr}$  using

\begin{equation}
	\label{eqn_theta_update}
	\theta_{t+1} = \theta_t + \alpha_{lr} J(\pi_\theta)\Bigr|_{\theta_t}
\end{equation}

Based on the \textit{policy gradient theorem} \cite{NIPS1999_1713}, the gradient of Equation \ref{eqn_expected_return} is given by

\begin{equation}
	\label{eqn_policy_gradient}
	\nabla_\theta J(\pi_\theta) = \mathbb{E}_{\tau \sim \pi_\theta} [\sum_{t=1}^T[\nabla_\theta \log \pi_\theta (a_t \mid s_t)] R(\tau)]
\end{equation}

As the left-hand side of Equation \ref{eqn_policy_gradient} is an expectation, it can be approximated by Monte-Carlo sampling using a set of trajectories. It is common to use the \textit{advantage function} $A(s,a) = Q(s,a) - V(s)$ instead of the discounted reward to reduce the variability (the advantage function can be estimated with methods such as Generalized Advantage Estimator (GAE) \cite{schulman_high-dimensional_2015}). Given a sampled set of trajectories $T = \tau_1, \dots, \tau_T$, each with observed states $s_0, \dots, s_j$, the weights $\phi$ of the value function approximation are updated by

\begin{equation}
	\phi_{k+1} = \arg\min_\phi \frac{1}{k |T|} \sum_{\tau \in T} \sum_{t=0}^T (\hat{V}_\pi(s \mid \phi) - R_t)^2
\end{equation}
being $R_t$ the sum of cumulative (discounted) rewards in each trajectory, as defined in Eqn. \ref{eqn_cumulative_rewards}.

A drawback of policy gradient algorithms is that they are prone to numerical instabilities in the policy as they are iteratively updated via a stochastic gradient. A recent algorithm that successfully addresses this issue is Proximal Policy Optimization (PPO) \cite{schulman_proximal_2017}, which is an on-policy algorithm that updates the policy $\pi_{\theta}$, by limiting (usually, by clipping) how much the weights can be changed when performing gradient ascent updates, making it a solid state-of-art algorithm.

\subsection{Behavioral cloning} 
\label{sub:behavioral_cloning}

An alternative way of learning behavioral policies is from a policy made from demonstrations $\tilde{\pi}$. This \textit{demonstration policy} consists of trajectories, i.e., $\tilde{\pi} = \{\tau_1, \dots, \tau_p\}$, each with state-action pairs $(s_0, a_0), \dots, (s_k, a_k)$, and the problem of learning an approximated policy $\hat{\pi}$ from them is called \textit{imitation learning} \cite{argall_survey_2009}. These demonstrations can originate from several sources. For instance, they can be manually generated by domain experts, generated by a partially trained reinforcement learning agent, or by transfer learning from similar tasks.

In particular, \textit{behavioral cloning} (BC) is a simple imitation learning algorithm that consists of learning $\hat{\pi}$ in a supervised learning setting, where a model such as a support vector machine (SVM) or neural network is trained to predict output actions given input states, typically by learning to minimize a function $\ell(\hat{a}, \tilde{a})$, that computes the loss of selecting action $\hat{a}$ instead of the correct action $\tilde{a}$. As in supervised learning, the demonstrations are usually split in training and validation sets to reduce overfitting. The loss function $\ell$ may take several forms; for instance, for continuous actions it can compute a similarity norm between two actions, whereas for discrete actions it can consist in the cross-entropy loss.

The main advantage of BC is related to its efficiency (a small set of demonstrations can be used with good results), and the velocity to train a learning agent. However, one of the drawbacks of learning policies solely with BC is that both the distribution of the demonstrations and the distribution of the $(s,a)$ that the agent will know after training are assumed to be IID \cite{ross_efficient_2010}, when they are instead dependent on past actions and states, and therefore need to be sampled by directly interacting with the environment. Such limitations are particularly significant in cases when an agent must take action at states that are sensibly different from the observed in the training set, which expose it to actions that are not only suboptimal, but also risky, resulting in catastrophic outcomes for certain environments. However, the simplicity of BC, when combined with deep reinforcement learning algorithms, can provide a fast and efficient way to pre-train policies in an RL setting. This gave rise to considerable successes. For example, by means of BC, very good results and efficient policy learning were achieved in Atari games \cite{hester_deep_2017}, in environments consisting of simulated human-like hands that manipulate objects \cite{rajeswaran_learning_2018}, and in a simulated quadcopter vehicle \cite{goecks_integrating_2020}.

\subsection{Bayesian optimization} 
\label{sub:bayesian_optimization}

Bayesian optimization \cite{mockus_application_1978,shahriari_taking_2016} is a black-box optimization method that aims to find the global maximum of an unknown function $f$, i.e., $\arg\max_X f(X), X \in \mathbb{R}^d$, by selectively sampling $X$ and optimizing a cheaper surrogate function designed to balance exploration and exploitation using an existing data set. For each new point $X$, the maximization of $f(X)$ is performed by taking a Bayesian approach over $f$, that is, assuming that 1) $f$ follows a prior probabilistic distribution, and 2) using Bayes' theorem for computing the posterior of $f$ given $D_n$, the set of $n$ pairs of queries and corresponding outputs.

A common approach involves the assumption that the noisy samples $y_{X_1}, \dots, y_{X_n}$ of the unknown objective function values, $f(X_1), \dots, f(X_n)$, are normally distributed, i.e., $f(X_1), \dots, f(X_n) \sim Normal(m,K)$, where:

\begin{itemize}
	\item $m = (\mu_0(X_1), \dots, \mu_0(X_n))$, being $\mu_0(X)$ the prior mean function, typically $\mu_0(.) = \mu_0$ (i.e., set to a constant prior mean). The prior mean represents the prior belief about the behavior of the objective function $f$, before any output is observed, and it can be obtained by inferring it from the data, or by expert knowledge (if it is available) \cite{shahriari_taking_2016}.
	\item $K$ is the covariance matrix that denotes the covariance $k(X_i, X_j)$ of each point $X_i$ with each other $X_j$. The covariance function $k(.,.)$ is defined by a covariance kernel that determines how two points will be influenced between each other, e.g., by a Matérn kernel \cite{shahriari_taking_2016}. Kernels usually have smoothness hyper-parameters that determine smoothness and amplitude of the samples. Such kernel hyper-parameters are usually updated after each iteration, to automatically adjust to the observed data (see Shahriari et al. (2016) \cite{shahriari_taking_2016} for more details about how this is done).
\end{itemize}

Under this assumption, the statistical model is a Gaussian process (GP) \cite{rasmussen_gaussian_2008} and their Bayesian treatment has a closed form based on the \textit{kernel trick} \cite{rasmussen_gaussian_2008}. This allows to use GP as a regression model to predict both the mean $\mu$ an uncertainty $\sigma^2$ of a new point $X_{n+1}$, taking into account previous data $D_n = {(X_1, y_{X_1}), \dots (X_n, y_{X_n})}$. This prediction is given by

\begin{eqnarray}
	\label{eqn_pred_mean}
	\mu_{n+1}(X) &=& \mu_0(X) + k(X)^T K^{-1} (Y - m) \\
	\label{eqn_pred_variance}
	\sigma^2_{n+1}(X) &=& k(X, X) - k(X)^T K^{-1} k(X)
\end{eqnarray}
where $Y = (y_{X_1}, \dots, y_{X_n})$, and $k(X)$ denotes the covariance between the new point $X_{n+1}$ and the previous points $X_1, \dots, X_n$ in the data set.

The parameters $\mu_{n+1}$ and $\sigma^2_{n+1}$ are used in order to obtain a new point for next query that, according to the information from previous points, maximizes the chances of encountering the true optimum of $f$. As the latter is unknown, the point that is commonly used as a reference is the point that yielded the current maximum, $X^+$. The next query point of $f$ is obtained by maximizing a function $\alpha(X^d) \to \mathbb{R}$ that is cheap to optimize (in comparison to $f$) called \textit{acquisition function}, which uses the predicted mean and variance from the statistical GP model to compare against $X^+$, and to determine the uncertainty of the prediction \cite{shahriari_taking_2016}.

\begin{algorithm}
	
\caption{Bayesian optimization}
\label{alg:bo_algorithm}
\begin{algorithmic}
	
\Input
\State prior mean $\mu_0$, covariance kernel function $k(.,.)$ acquisition function $\alpha(.)$, unknown black-box function $f$
\EndInput

\For{$\text{evaluation} = 1$ to N evaluations}
	
	\State Obtain $X_{n+1}$ by optimizing an acquisition function $\alpha(X)$ using the predicted $\mu_{n+1}$ and $\sigma_{n+1}$ from a statistical model (e.g., Gaussian Process)
	
	\State Query the objective function $f$ at the point $X_{n+1}$
	
	\State Add the point and result $y_{X_{n+1}}$ to $D_{n+1} = \{(X_1, y_{X_{1}}), ..., (X_{n+1}, y_{X_{n+1}})\}$
	
	\State Update the statistical model (e.g., Gaussian process)
\EndFor

\Output

$\arg\max_X, y_X$

\EndOutput
\end{algorithmic}
\end{algorithm}

The pseudo-code for Bayesian optimization is described in Algorithm \ref{alg:bo_algorithm}. It consists of a number of $N$ queries to be made to the objective function $f$, that are performed in an iterative way in order to determine the next point $X_{n+1}$. As $f$ is assumed to be expensive to calculate, the main intention of the algorithm is that each query point is chosen based on the information of previous queries. In order to define the next point $X_{n+1}$ that will be queried, the most common acquisition function used is the \textit{expected improvement} (EI) function. It estimates the probability that a test point $X$ might yield an improvement over the best possible outcome $f^*$, which is pondered by the amount of improvement expected for such a test point. Under the normal distribution assumption, this acquisition function is given by

\begin{eqnarray}
	\label{eqn:ei}
	\alpha_{EI} (X) &=& \mathbb{E}[f(X) - f^*]  P[f(X) > f^*]  \\
	&=& (\mu_n(X) - f^*) \Phi(Z) + \sigma_n(X) \phi(Z) \nonumber
\end{eqnarray}

where $\Phi(.)$ and $\phi(.)$ are the standard Gaussian cumulative and density distribution functions, respectively. As it is unknown, a common criterion is to use the current maximum found according to the selected metric, i.e., $f^* = f(X^+)$, being $X^+$ the current vector that obtains the maximum $y$. Finally, $Z = (\mu_{n}(X) - \tau)/\sigma_n(X)$.

Typically, the maximum of Eqn. \ref{eqn:ei} is obtained in practice by sampling a batch of points $X$ selected at random, or by resorting to sampling methods such as \textit{Latin Hypercube Sampling} (LHS), that divides the search space in a grid, and then sample points in such a way that each selected sample is the only sample in the corresponding row and column.

This Bayesian optimization approach is commonly used to optimize the hyper-parameters $\vartheta$ of machine learning algorithms \cite{snoek_practical_2012}, where each algorithm execution is taken as a black-box function $f(\vartheta)$, where $f$ is a previously determined measure of how well the agent has learned. For example, in the context of this work, $f$ can represent the effectiveness of a reinforcement learning algorithm for finding a policy to solve a certain task. On the other hand, $\vartheta$ is a vector that contains the elements that define the configuration that is being optimized in the algorithm, that can be composed of real valued, discrete or combinatorial elements (in such case a Bayesian optimization approach such as the proposed by Baptista et al. (2018) \cite{baptista_bayesian_2018} is recommended).
Aside from the method presented in this Section, other variants of Bayesian optimization are also used to optimize hyper-parameters. One common common variant was introduced by Hutter et al. (2011) \mbox{\cite{hutter_sequential_2011}}, which proposes random forest regression as a surrogate model.
Another variation that is used in several implementations is the Tree-structured Parzen Estimator (TPE) approach \mbox{\cite{bergstra_algorithms_2011}}, which estimates the expected improvement by generating separately $P(y)$ and $P(X \mid y)$, the latter by using two non-parametric densities, instead of directly modeling $P(y \mid X)$ with Gaussian process regression as in Eqn. \mbox{\ref{eqn_pred_mean}}.


\section{Related work in hyper-parameter optimization} 
\label{sub:related_work_in_hyper_parameter_optimization}

Different approaches have been applied to hyper-parameter setting optimization in machine learning algorithms, related to supervisory learning tasks. An often-used methodology is the expensive \textit{grid search}, which mainly consists in an exhaustive search over the hyper-parameter space. To circumvent the computational cost of grid search, the less costly alternative of \textit{random search} has also been tried \cite{bergstra_random_2012}. The latter revolves around a random-guided search over a hyper-sphere, which only takes into account the information of the best point. By resorting to pure luck in the search for a "good" setting of algorithm hyper-parameters, random search does not take full advantage of information gained from the sequence of past queries.
Other approaches involve evolutionary methods, such as the Covariance Matrix Adaptation Evolution Strategy (CMA-ES) \mbox{\cite{hansen_completely_2001, loshchilov_cma-es_2016}}, which consists in iterations where candidate solutions are sampled from a multivariate normal distribution; such samples are evaluated and used to update the covariance matrix of the distribution, in order to maximize the likelihood of successful solution points.
However, as it is common with evolutionary approaches, they require a considerable computational budget to sample and update many candidate solutions \mbox{\cite{loshchilov_cma-es_2016}}.

As an improvement over uninformative methods like random search, or exhaustive methods such as grid search, the Bayesian optimization of hyper-parameters implements a sequential, iterative optimization of a black box function (in this context, a performance function of the hyper-parameters), by optimizing a cheaper function that takes into account past queries to decide the next setting of algorithm hyper-parameters.
Given its performance and efficiency to tune hyper-parameters, Bayesian optimization gave rise to a series of novel frameworks for auto-tuning the hyper-parameters in machine learning algorithms, such as the Sequential Model-based Algorithm Configuration (SMAC) (based on Hutter et al. (2011) \cite{hutter_sequential_2011}), Auto-WEKA \cite{thornton_auto-weka_2013}, MOE \cite{clark_moe:_2014}, Hyperopt \cite{bergstra_hyperopt:_2013}, Spearmint (based on Bergstra and Bengio (2012) \cite{bergstra_random_2012}), and more recently, the combination of traditional Bayesian optimization with the Hyperband strategy \cite{li_hyperband_2017}, resulting in BOHB \cite{falkner_bohb_2018}, and the integration in Optuna \cite{akiba_optuna_2019}. However, none of these approaches are suited for RL, since they separate data generation from the learning task, hence it is very inefficient to assume that the algorithm hyper-parameters can be optimized without affecting the information content of the generated data set to approximate the optimal policy. 

In particular, regarding reinforcement learning algorithms, several approaches have been taken with respect to optimizing their hyper-parameters. Some studies focus separately on each hyper-parameter. For instance, an approach to optimize the trace hyper-parameter $\lambda$ by minimizing the variance error of the estimation of the value without incurring in a high bias error is presented in White and White (2016) \cite{white_greedy_2016}. A way to adapt the step size $\alpha_{lr}$, according to derived upper and lower bounds where the learning performed with function approximation will not diverge, is discussed in Dabney and Barto (2016) \cite{dabney_adaptive_2012}.

Other studies, on the other hand, focused on resorting to Bayesian optimization to optimize several RL hyper-parameters at the same time.
In this setting, a preceding work is Barsce et al. (2017) \cite{barsce_towards_2017}, where a Bayesian optimization framework was proposed to optimize RL hyper-parameters. However, in such work, Bayesian optimization and the RL algorithm are decoupled in such a way that the meta-learning does not make specific assumptions with respect to an RL algorithm, making the method also inefficient because the learned tuples of experience $(s,a,r,s')$ are all aggregated in the metric selected for the objective function and, therefore, cannot be used directly to improve the meta-learning layer.
On the other hand, Bayesian optimization with Tree-structured Parzen Estimators (TPE), in combination with Hyperband \mbox{\cite{falkner_bohb_2018}}, was also validated in Falkner et al. (2018) \mbox{\cite{falkner_bohb_2018}}, as a suitable BO alternative to optimizing RL parameters, by optimizing eight hyper-parameters in a Cartpole swing-up task.
Moreover, Chen et al. (2018) \mbox{\cite{chen_bayesian_2018}} employed a sophisticated version of Bayesian optimization with self-play, to improve the hyper-parameters of AlphaGo \mbox{\cite{silver_mastering_2016}}.
On the other hand, Liessener et al. (2019) \mbox{\cite{liessner_hyperparameter_2019}} used Gaussian process and random forest variants of BO in order to perform hyper-parameter tuning of a deep RL agent immersed in a vehicle energy management problem.
In another work, Young et al. (2020) \cite{young_distributed_2020} proposed a framework that runs many Gaussian processes in parallel, where each one runs in a fraction of a divided hyper-parameter search space, in order to optimize three Atari games; however, this approach is very demanding computationally, and hardware resources are a bottleneck when exploring asynchronously.

A recent direction of hyper-parameter tuning in RL involves automatically tuning many of the learning algorithm hyper-parameters in an online setting (i.e. changing its values as learning occurs), and proposes an alternative approach, which can be combined with sequential models to learn good hyper-parameter initializations that are later adjusted online.
This is consistent with the notion that changing the values of hyper-parameters, e.g. gradually increasing the discount factor such as in Francois-Lavet et al. (2016) \cite{francois-lavet_how_2016}, may be beneficial during the different stages of training of the agent.
One of this online approaches involves updating the differentiable hyper-parameters $\eta$, by taking the gradient of an also differentiable actor-critic loss function with respect to $\eta$ (this is known as a meta-gradient, as such loss also involves taking a gradient with respect to the algorithm parameters $\theta$).
The first work to apply meta-gradient in RL for hyper-parameter tuning was proposed by Xu et al. (2018) \cite{xu_meta-gradient_2018}, which resorts to meta-gradients to optimize the hyper-parameters $\gamma$ and $\lambda$ of an IMPALA agent, and it is further improved in Zahavy et al. (2020) \cite{zahavy_self-tuning_2020-1}, by augmenting the number of self-tuning hyper-parameters in an extended version of IMPALA.
Meta-gradients have also been used for optimizing high-level hyper-parameters, such as in auxiliary tasks \cite{NEURIPS2019_10ff0b5e}, the return function \cite{wang_beyond_2020}, or the actor-critic loss \cite{NEURIPS2020_cceff8fa}.
Another online tuning approach was proposed in Tang et al. (2020) \cite{tang_online_2020}, that involves performing agent rollouts with hyper-parameters sampled from a distribution, and, based on the collected experience, update the means of the hyper-parameters using an estimator of their gradients via Evolution Strategies (ES) \cite{salimans_evolution_2017}.
This ES approximation makes room for self-adjustment of hyper-parameters that are not differentiable, thus extending to off-policy algorithms.

Other optimization approaches include optimizing RL via population based methods, that involves training (usually in parallel and with high computational capabilities) a population of candidate solutions, selecting the best and including them in the population of the next generation.
For instance, in Jaderberg et al. (2017) \cite{jaderberg_population_2017}, a population of agents is trained using a distributed Asynchronous Actor-Critic (A3C) setting \cite{mnih_asynchronous_2016}, that combines both the optimization of parameters and hyper-parameters. For the latter, each new population is chosen based on the performance of its members, whereas small perturbations to the hyper-parameters are also applied to search for new solutions.
Other recent population-based approach is proposed by Franke et al. (2020) \cite{franke_sample-efficient_2020}, which focuses on off-policy algorithms, as it uses a shared-experience buffer \cite{schmitt_off-policy_2019} to train the members of a population, whereas the members of the next one are chosen via tournament selection with elitism, applying mutations in order to add exploration.
The shared experience buffer facilitates agent training from experience replay, thus reducing the interactions needed with the actual environment.

Finally, there are approaches that seek to address hyper-parameter optimization in different settings, such as Paine et al. (2020) \cite{paine_hyperparameter_2020}, which addresses hyper-parameter selection in off-line RL, or Zhang et al. (2021) \cite{zhang_importance_2021}, where the optimization of hyper-parameters in model-based RL is studied, and in Bouneffouf and Claeys (2021) \cite{bouneffouf_online_2021}, which focus on optimizing exploration parameters in contextual bandits.

As presented in the next Section, the approach proposed in this work focuses on making the most of the gathered experience, aiming to increase the efficiency of the hyper-parameter optimization process while generating satisfactory policies.
The focus of our proposal is quite different from general optimization methods which are typically used to optimize reinforcement learning models, and methods used to tune RL algorithms such as population-based methods, which demand high amounts of computational time to train the populations of candidate solutions.
Other recent approaches, such as meta-gradient, also differ from the focus of the current work; however, they are similar in that both use the collected experiences to perform an efficient update of the hyper-parameters, and the online setting proposed in meta-gradient approaches can be combined with the sequential optimization applied in this work, for instance to use the latter to determine the initial hyper-parameters of the former.

%


\section{Bayesian optimization of an RL agent}
\label{sec:proposed_method}

The proposed method consists of an architecture that uses Bayesian optimization to facilitate autonomous tuning of all the hyper-parameters of an RL agent. The reasoning behind integrating Bayesian optimization with an RL algorithm is that using a black-box approach is well suited for an expensive optimization task such as RL, by making the most of past queries in order to maximize the gain of selecting the next query of hyper-parameter setting through the acquisition function. As it employs Bayesian optimization for optimizing RL algorithms, this architecture has its foundations in RLOpt \cite{barsce_towards_2017}.

The novel aspect of the proposed approach, named \textit{RLOpt-BC}, is that it is adapted for an active learning setting in which an agent interacts with an environment to generate maximally informative experience to learn a way of behaving. To this aim, an RL-variant of the expected improvement criterion that takes advantage of pre-training from demonstrations is used. In this setting, the objective function selected is a function $f: \mathbb{R}^d \to \mathbb{R}$ that measures the overall performance of the agent, given an environment and a reinforcement-learning agent with a hyper-parameter vector $\vartheta = (\vartheta_1, \dots, \vartheta_d)$, each $\vartheta_i$ being a specific hyper-parameter (e.g., the learning rate hyper-parameter, $\alpha_{lr}$). As the latent distribution of the objective function $f$ is unknown, the prior assumption is that it can be functionally approximated by a Gaussian process, whose means and variances are defined over the  state-action space.

In order to measure the overall performance, the learning agent is set to run for a number of episodes, where it resorts to a learning algorithm with a fixed vector of hyper-parameters.
In this work, this set of episodes is referred to as a \textit{meta-episode}, to highlight that Bayesian optimization is applied at a higher level of abstraction where the hyper-parameters are systematically modified in order to maximize the output value of the metric for the function $f$, closely related to speeding up the learning curve.

To estimate the value of $f(\vartheta)$ in a given meta-episode, an RL agent is instantiated with the configuration $\vartheta$ in a defined environment, and set to learn a policy to maximize its expected reward for a fixed number of episodes (or time steps, when the task is continuous). Based on the generated data (which consists of state transitions, actions taken and rewards obtained), a performance value for the resulting policy is calculated and stored, whereas the learned policy is erased, so the agent can be instantiated again with other combination $\vartheta$  of hyper-parameters, and start a new meta-episode. The metric chosen to account for the agent performance in each meta-episode depends on the particularities of the learning task. For example, the value of $f(\vartheta)$ can be the average of the rewards gathered in the meta-episode using a given $\vartheta$, the maximum cumulative reward for a number of time steps, or simply the maximum individual reward received.

Taking into account that agent-environment interactions are expensive and often risky, it is important to bias learning using suboptimal policies learned in past meta-episodes. To this aim, before beginning each new meta-episode, the RL agent is pre-trained with a subset of demonstrations from the best performing policy found so far. Whenever a tuple of experience is used to pre-train the agent, its impact on the learning process will be determined based on the current setting of hyper-parameters. As a result, each time the agent begins a meta-episode, the policy is not learned from scratch, but it is biased using knowledge from previous experiences, which speed up the learning curve and favor fast detecting whether a setting of hyper-parameters is suitable or not for the task the agent is trying to solve.

For past experience to be used in pre-training the learning agent with BC, each time a new maximum $f(\vartheta)$ is found, a number of demonstration trajectories are generated by recording data from testing episodes using the corresponding policy, and stored in a set $\psi$, replacing previous demonstrations with new trajectories using the current best policy. This set of trajectories is used to pre-train both the simulations of the acquisition function, and the queries of the objective function $f$. It is worth noting that the information content of these tuples is independent from the hyper-parameters used to generate them; however, bias does exist in the set $\psi$, because highly informative tuples were found by resorting to a certain combination of hyper-parameters.

The proposed approach involves extending the standard expected improvement acquisition function to a form of simulation-based learning of the effect of hyper-parameters setting. This is performed in two steps to estimate the optimum by trying a sequence of hyper-parameters: In the first, the standard expected improvement acquisition function, based on prior queries $D_n = \{(\vartheta_i, y_{{\vartheta}_i})\}_n$, is used to obtain the top $m$ best candidate points, $\vartheta_{1}, \dots, \vartheta_{m}$, from a sampled batch of points (e.g., chosen at random or by LHS). In the second step, $m$ policies, one for each candidate point, are pre-trained from a subset of the demonstration and then set to run for a small number of episodes. When they finish, the expected improvement of each $m$ is calculated from the received rewards, given their empirical mean $\hat{\mu}$ and standard deviation $\hat{\sigma}$. The point that reached the maximum EI is selected as the new configuration $\vartheta$ to query the performance function $f$ in the next meta-episode.

The extension of the acquisition function is described in Algorithm \ref{alg:ei_rl}. In the latter, $\{\tau_1, \dots, \tau_i\}$ is a pre-selected subset of trajectories from the set $\psi$, sampled with a random uniform probability (although this subset could also be sampled according to some other information content metric, such as the total reward obtained by the agent in the episode where the tuple was seen).

\begin{algorithm}
\caption{EI extended for computing maximum expected reward $\sigma_{EI-BC}(\vartheta)$}
\label{alg:ei_rl}
\begin{algorithmic}

\Input
\State set of demonstrations $\psi$, size of sampled demonstrations $i$, number of rollout policies $m$, prior mean $\mu_0$, covariance kernel function $k(.,.)$, RL environment
\EndInput

\State Obtain the best $\vartheta_{1}, \dots, \vartheta_{m}$ points of the EI acquisition function (Eqn. \ref{eqn:ei}), by sampling a batch of points (e.g., via random sampling) 

\State Sample a subset of demonstrations $\{\tau_1, \dots, \tau_i\} \subseteq \psi$

\State Learn $m$ policies by behavioral cloning with past experience $\{\tau_1, \dots, \tau_i\}$

\For{p=1 to $m$-th policy}
	\State Run the agent in the environment for $e$ episodes with learned policy $\pi_{\vartheta}$. Obtain rewards $r_1, \dots, r_e$
	
	\State Calculate the empirical mean $\hat{\mu}$ and standard deviation $\hat{\sigma}$, given episode rewards $r_1, \dots, r_e$
	
	\State Calculate expected improvement for the policy, $ei_p$, given $\hat{\mu}$ and $\hat{\sigma}$, according to Eqn. \ref{eqn:ei} 
	
\EndFor
	
\Output

\State $\arg\max_\vartheta, \max(ei_1, \dots, ei_m)$
\EndOutput
	
\end{algorithmic}
\end{algorithm}

The proposed variant of expected improvement allows the agent to determine how each specific vector $\vartheta$  of hyper-parameters performs when compared with the current best, $\vartheta^+$. This is done by taking into account how the agent learns a policy, given a vector of hyper-parameters and a sample of past experiences within a limited budget of simulated interactions with its environment. Under this setting, past outputs  $y_{\vartheta}$ are also used in an indirect way to calculate the empirical mean and variance. Therefore, both previous pairs $(\vartheta, y_{\vartheta})$, and highly informative data from past episodes using the best policy are used. As a result, purposefully biasing policy learning through hyper-parameter setting is a key point of the algorithm. Thus, by integrating Algorithm \ref{alg:ei_rl} and behavioral cloning pre-training, Algorithm \ref{alg:bo_with_rl_acq} is proposed as a novel two-level reinforcement learning algorithm equipped for autonomous setting of its own hyper-parameters using Bayesian optimization.

\begin{algorithm}
\caption{Bayesian optimization extended for reinforcement learning}
\label{alg:bo_with_rl_acq}
\begin{algorithmic}

\Input
\State $n$ meta-episodes, hyper-parameters to optimize $\vartheta$, RL environment, size of demonstrations $|\psi|$, size of sampled demonstrations $i$, number of pretrain episodes, episodes to run per meta-episode
\EndInput

\For{$\text{meta-episode} = 1$ to $n$ meta-episodes}
	
	\State Obtain $\vartheta_{n+1}$ by optimizing $\sigma_{EI-BC}$ (Algorithm \ref{alg:ei_rl})
	
	\State Initialize a new reinforcement learning agent with hyper-parameters $\vartheta_{n+1}$ and no prior learning 
	
	\State Sample a trajectory subset $\{\tau_1, \dots, \tau_i\} \subseteq \psi$ 
	
	\State Pretrain policy $\pi_{\vartheta_{n+1}}$ with BC using $\{\tau_1, \dots, \tau_i\}$ 
	
	\State Obtain $y_{\vartheta_{n+1}}$ by sampling the objective function $f$, starting from pretrained policy $\pi_{\vartheta_{n+1}}$ 
	
	\If{$y_{\vartheta_{n+1}}$ is the new maximum}
		
		\State With trained policy $\pi_{\vartheta_{n+1}}$, generate a new set of demonstration trajectories $\psi_{new}$ 
		$\psi \gets \psi_{new}$
	
	\EndIf
	
	\State Add $(\vartheta_{n+1}, y_{\vartheta_{n+1}})$ to the data $D_n$ 
	
	\State Update the statistical model

\EndFor

\Output

\State $\arg \max_{\vartheta^+}, y_{\vartheta^+}$

\EndOutput

\end{algorithmic}
\end{algorithm}

In Algorithm \ref{alg:bo_with_rl_acq}, the calculated output $y_{\vartheta}$ for the performance metric $f$ is a measure of how well the agent performs in the environment based on the received rewards, given that the learning policy was generated using the hyper-parameter configuration $\vartheta$. However, the particular aspect of the proposed approach is that, as each sampling of $f(\vartheta)$ corresponds to a realization of a reinforcement learning agent, its value accounts for previous knowledge gathered from state transitions experienced when using $\vartheta$, and thus the execution is influenced by the previous acquired experience tagged by different hyper-parameter combinations, as it happens in Algorithm \ref{alg:ei_rl}. Because of that, the queried points demand lower computational costs (because the policy learning is conveniently biased  using previous experience), thus significantly reducing the inherent costs of a reinforcement learning agent starting afresh.

From the perspective of a reinforcement learning agent that is optimizing the function $f(\vartheta)$, both algorithms presented above combine meta-learning of the hyper-parameters with learning how to solve the task by interacting with its environment at the lower level. In other words, the overall aim of Algorithm \ref{alg:bo_with_rl_acq} is \textit{learning to learn} how to solve a reinforcement learning problem. In this meta-learning setting addressed here by BO, proposed algorithms define which are the hyper-parameters that speed up the learning curve, which in turn provide highly informative state transition according to the selected configuration, and whose learning is also used to better define the next setting of hyper-parameters, and so on. In this sense, there is another aspect worth mentioning: both ground-level learning and meta-learning share the traces of experience $\psi$, which means that meta-learning also influences the learning from interaction by changing hyper-parameters to find more rapidly the optimal policy. On the other hand, the meta-learning that the agent performs also influences future learning to a significant extent, because tuples experienced by the agent by using a given configuration $\vartheta$  will be used to initialize the policies of future agent realizations, and thus each time a new learning agent is created for a task, it can take advantage of knowledge generated by previous agents.

\section{Computational experiments}
\label{sec:experimental}

In this Section, the method proposed in Section \ref{sec:proposed_method} is mainly validated in several robotic simulation environments from the PyBullet framework \cite{coumans2019} and in BSuite environments \cite{osband_behaviour_2019} as an additional validation, where the Stable-baselines framework \cite{stable-baselines} was used to run the RL agents that had their hyper-parameters optimized. In this Section, the run of a determined set of meta-episodes is called an \textit{execution}, and each execution consists of 10 meta-episodes. All the experiments were run in a computational environment consisting of an AMD Ryzen 5 3600X processor with a base frequency of 3.80GHz and 16 GB of RAM. Details of each experimental setting and the corresponding learning curves are presented in the following subsections.

\subsection{Validation in PyBullet robotic simulation environments}
\label{sec:pybullet_simulations}

\begin{figure*}
	\centering
	\begin{tabular}{ll}
		\includegraphics[height=0.25\textwidth, trim={15 15 18 15}, clip]{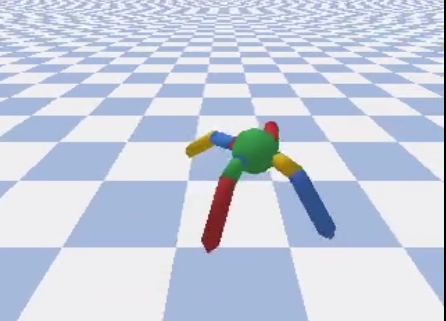}
		&
		\includegraphics[height=0.25\textwidth, trim={15 15 24 15}, clip]{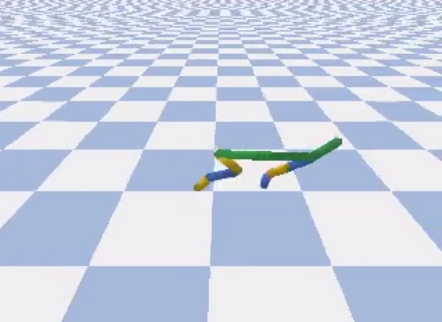} \\
		
		\includegraphics[height=0.25\textwidth, trim={15 15 15 15}, clip]
		{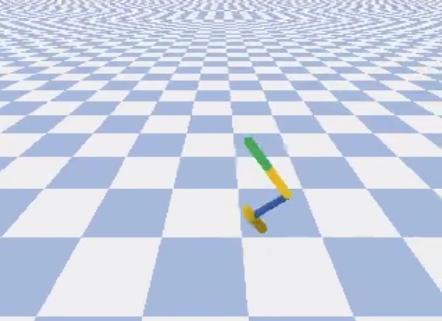}
		&
		\includegraphics[height=0.25\textwidth, trim={15 15 15 15}, clip]
		{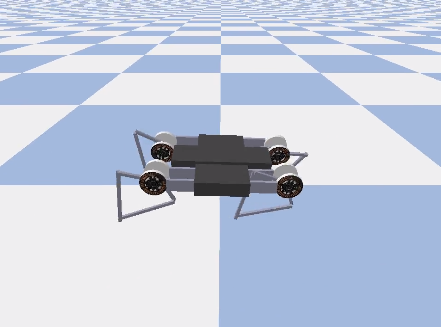}
	\end{tabular}
	
	\centering
	\includegraphics[height=0.25\textwidth, trim={0 0 0 15}, clip]
	{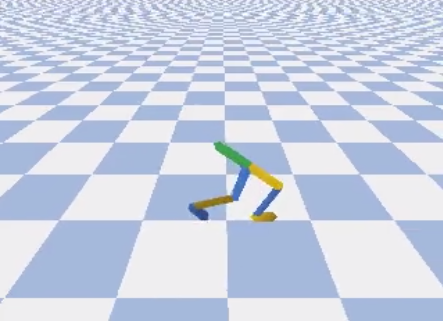}
	
	\caption{Different environments used to validate the methods. Up-left: Ant 3D robot. Up-right: Half-cheetah 2D robot. Mid-left: Hopper 2D robot. Mid-right: Minitaur 3D robot. Down: Walker 2D robot.}
	\label{fig:env_figures}
\end{figure*}

In this validation, each of the optimizers were run in five robotic simulation environments (depicted in Fig. \ref{fig:env_figures}), where a reinforcement learning agent must learn to control its joints to make them move and maximize the received reward, based on the distance travelled to the right. Each meta-episode consisted of 2 million time-steps. The following were the evaluated environments:

\begin{itemize}
	\item Ant Bullet env, which simulates an ant-like 3-Dimensional robot with 4 legs.
	\item Half-cheetah Bullet env, simulating a 2D 2-legged robot able to run.
	\item Hopper Bullet env, simulating a 1-leg 2D robot that hops.
	\item Minitaur Bullet env, which simulates a 3D quadruped robot, as described in Tan et al. (2018) \cite{tan_sim--real_2018}.
	\item Walker Bullet env, which simulates a walking 2D robot with 2 legs.
\end{itemize}
For more information regarding PyBullet environments, the reader is referred to the  \href{https://docs.google.com/document/d/10sXEhzFRSnvFcl3XxNGhnD4N2SedqwdAvK3dsihxVUA/edit?usp=sharing}{PyBullet Quickstart Guide}.

A total of 6 executions, each with a different random seed, was used to run each hyper-parameter optimization alternative in each environment. The selected algorithm was Proximal Policy Optimization (PPO), as it is a solid state-of-art algorithm with a clipping mechanism to avoid divergence when updating the policy. Different optimization approaches were compared for optimizing the objective function $f(\vartheta)$ throughout the learning curve. The optimizers compared were the following:

\begin{itemize}
	\item RLOpt-BC optimizer, which is the approach proposed in this work. A number of 10 top candidate points are selected each time from the expected improvement acquisition function, where each simulation in the $\sigma_{EI-BC}(\vartheta)$ was executed for 25000 time-steps. The size of the demonstration set $\psi$ was set to 5 episodes (this corresponds approximately to ~2-5\% of each execution episode).
	
	\item Bayesian Optimization with Gaussian process and the standard expected improvement function, resorting to the implementation of Barsce et al. (2017) \cite{barsce_towards_2017} (RLOpt).
	
	\item Bayesian optimization using Tree-structured Parzen Estimators (TPE) \cite{bergstra_algorithms_2011}.
	This optimizer was implemented with the Optuna backend \cite{akiba_optuna_2019}:
	as Optuna features pruning mechanisms that cut off unpromising runs (\textit{trials}, as they are called in Optuna) before they finish, instead of using a fixed set of meta-episodes, they were set to run for roughly the same number of time steps used by RLOpt-BC to run 10 meta-episodes. As the number of trials that are run in Optuna in such time is usually greater than 10, in order to compare its results with the other optimization algorithms, a meta-episode in this context consists of 1/10th of the trials run in Optuna. In the optimizer shown in the plots, the Successive Halving pruner \cite{jamieson_non-stochastic_2016} was used, as it performed better than no pruning (i.e., setting Optuna to run for 10 trials, without early termination of unpromising meta-episodes) or using the Hyperband pruner (as it was proposed in the BOHB variant \mbox{\cite{falkner_bohb_2018}}).
	
	\item Random Search \cite{bergstra_random_2012}, consisting of an optimizer that chooses random hyper-parameter combinations in a hyper-sphere, whose centroid is located in the point of the best maximum found so far, and replaced with the next optima as they are found.
	As an alternative to random search, the random sampler from Optuna \mbox{\cite{akiba_optuna_2019}}, which samples at random incorporating prunning capabilities, was also tried.
	However, it did not yield better results than random search, therefore, it was not included in the plots.
	
	\item Sequential Model-based Algorithm Configuration (SMAC) \cite{smac-2017}, which is originally based on Hutter et al. (2011) \cite{hutter_sequential_2011}, and implements Bayesian optimization using random forest as their underlying model used to estimate the mean and variance of candidate points. This optimizer was implemented using the SMAC3 backend, available at \url{https://github.com/automl/SMAC3}.
		
	\item Additional optimizers were also compared, but not included in the main plots for visual clarity, as their performance were not significant: CMA-ES {\cite{loshchilov_cma-es_2016}}, BOHB {\cite{falkner_bohb_2018}} and the Optuna random sampler.
	The details of such optimizers are presented in Appendix \mbox{\ref{appdx_additional_experiments}}.

\end{itemize}

Performance assessment of the trained policy was made for all optimizers by evaluating the policy for five episodes every 100000 time steps (i.e., agents were evaluated 20 times), and the metric to evaluate $f(\vartheta)$ was set as the best performance value found during the execution. This was done with the intention of valuing each policy by its best performing outcome, given that this set of environments can present some minor fluctuations during training. The optimized hyper-parameters and their corresponding ranges were the following:

\begin{itemize}
	\item Learning rate $\alpha_{lr} \in (1e^{-4}, 1e^{-3})$ of the Adam optimizer \cite{kingma_adam_2017}.
	\item Discount factor $\gamma \in (0.8, 0.9999)$.
	\item Clip range $\in (0.1, 0.5)$, which determines how much the new policy can differ from the old policy.
	\item Generalized advantage estimator (GAE) \cite{schulman_high-dimensional_2015} parameter $\lambda \in (0.85, 0.9999)$.
	\item Entropy coefficient $\in (0, 0.1)$, which determines the proportion of the entropy bonus in the loss function.
	\item Value function coefficient $(0.5, 1)$, which determines the weight of the value function estimation in the loss function.
\end{itemize}

\begin{figure*}[hb!]
	\centering
	\begin{tabular}{ll}
		\includegraphics[height=0.39\textwidth]{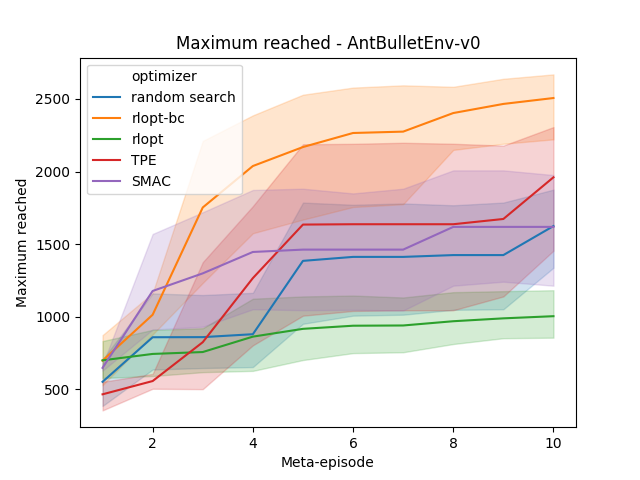}
		&
		\includegraphics[height=0.39\textwidth]{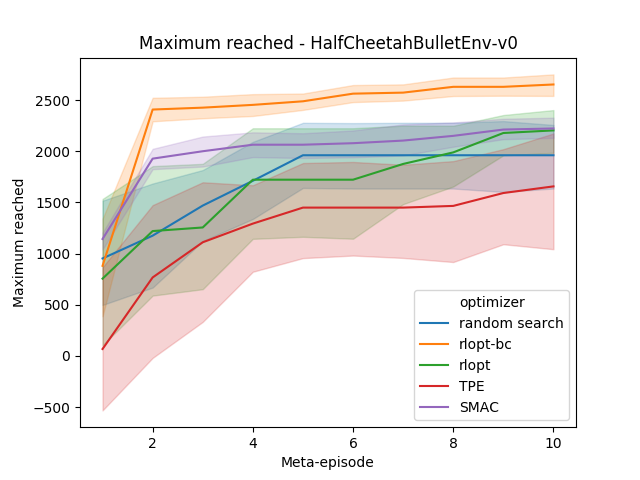} \\
		
		\includegraphics[height=0.39\textwidth]{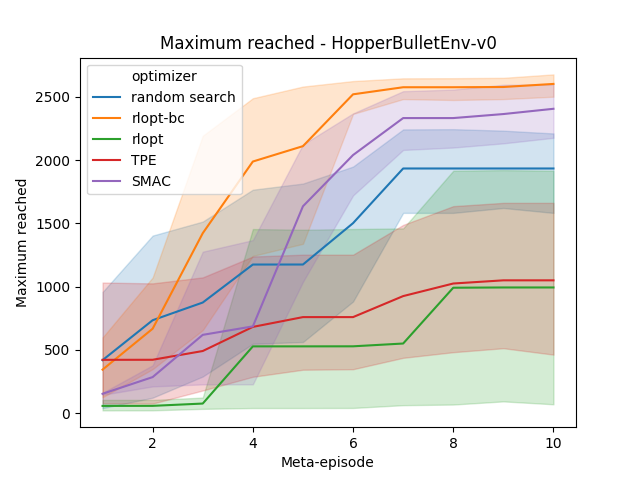}
		&
		\includegraphics[height=0.39\textwidth]{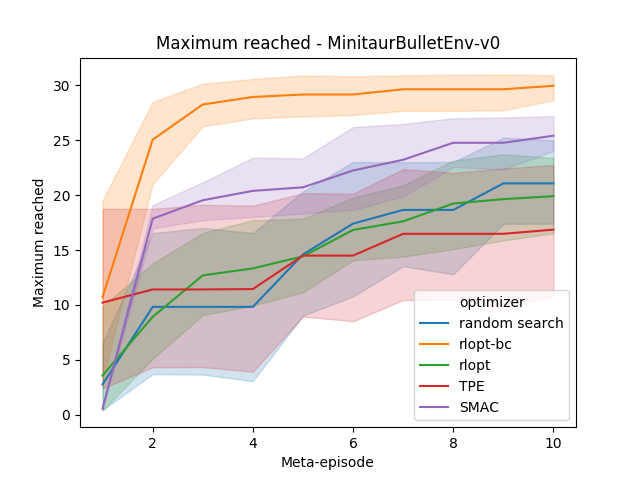}
		
	\end{tabular}
	
	\centering
	\includegraphics[height=0.39\textwidth]{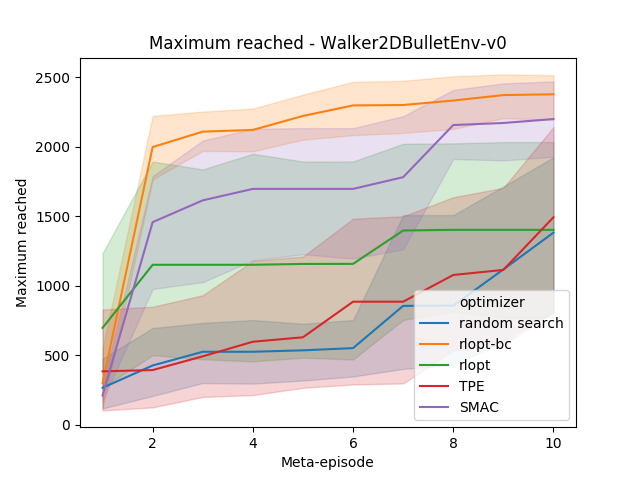}
	
	\caption{Comparison of the average maximum reached at each meta-episode for each of the optimizers, considering six different executions.}
	\label{fig:maximums_reached}
\end{figure*}

\begin{figure*}[hb!]
	\centering
	\begin{tabular}{ll}
		\includegraphics[height=0.39\textwidth]{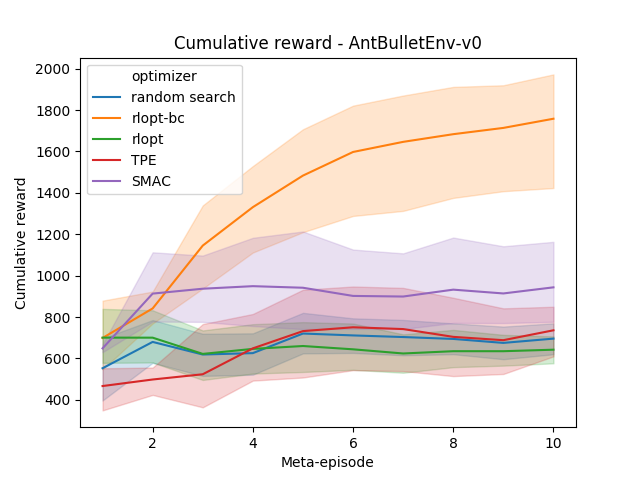}
		&
		\includegraphics[height=0.39\textwidth]{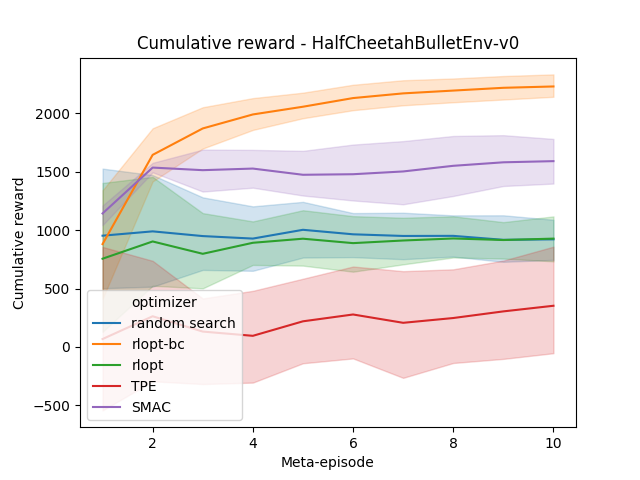} \\
		
		\includegraphics[height=0.39\textwidth]{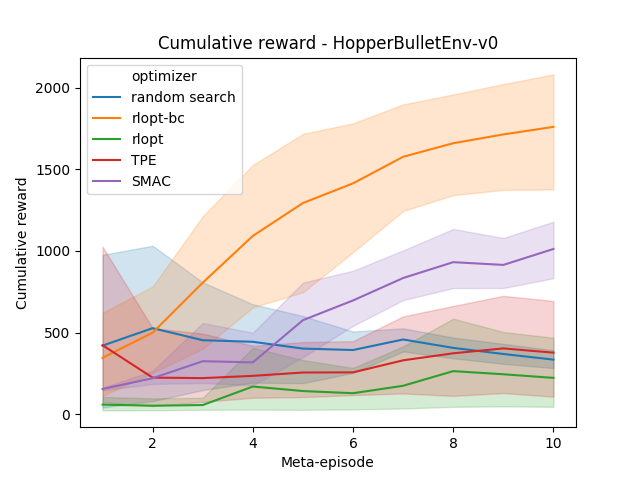}
		&
		\includegraphics[height=0.39\textwidth]{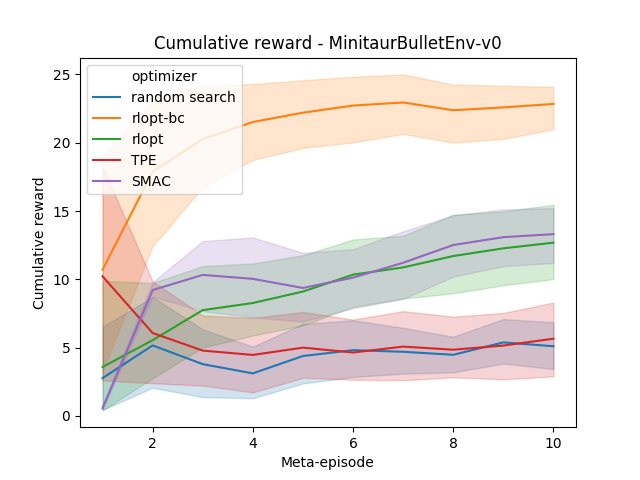}
		
	\end{tabular}
	
	\centering
	\includegraphics[height=0.39\textwidth]{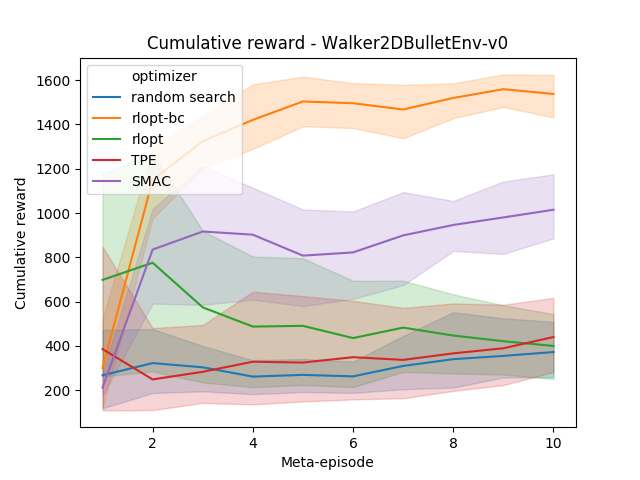}
	
	\caption{Comparison of the average cumulative rewards reached at each meta-episode for each of the optimizers, considering six different executions.}
	\label{fig:cumulative_rewards}
\end{figure*}

Results obtained are shown in terms of the maximum reached in Fig. \ref{fig:maximums_reached} and in terms of their average cumulative rewards in Fig. \ref{fig:cumulative_rewards}, in plots where the different optimizers are compared based on their corresponding maxima and rewards, respectively, across different meta-episodes. With respect to the maximum reached represented in Fig. \ref{fig:maximums_reached} for each optimizer, the tick lines and their nearby curves correspond to the average maximum found and the 95\% confidence interval for 6 executions with different random seeds. It can be appreciated that the proposed approach outperforms the other optimization alternatives by a considerable margin.
Regarding the average cumulative rewards in Fig. \ref{fig:cumulative_rewards}, it can be seen that the average rewards gathered by the proposed approach tends to increase from one meta-episode to another, being always higher than the average rewards for the other optimizers.
The average running times are detailed in Table \ref{tbl:running_times_pybullet}, where TPE times were marked with $^*$ to remark that it was set to run for that fixed amount of time, instead of a fixed set of meta-episodes, so that the Optuna backend can prune unpromising trials for the duration of the total optimization time.

\begin{figure*}[hb!]
	\centering
	\begin{tabular}{ll}
		\includegraphics[height=0.395\textwidth]{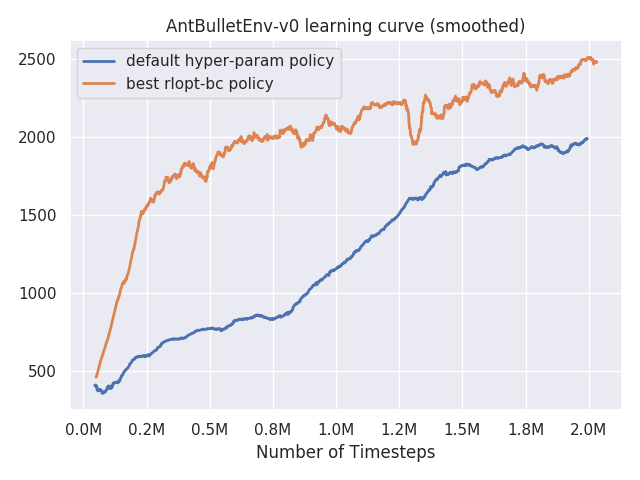}
		&
		\includegraphics[height=0.395\textwidth]{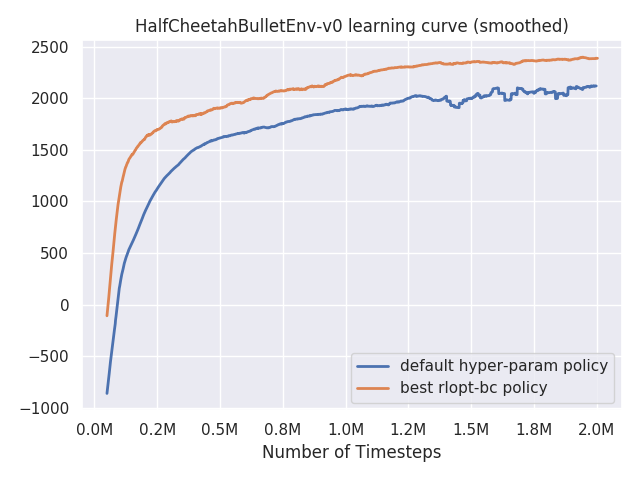} \\
		
		\includegraphics[height=0.395\textwidth]{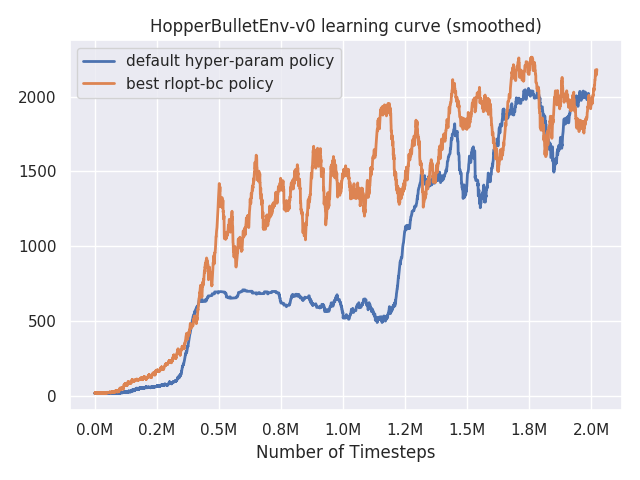}
		&
		\includegraphics[height=0.395\textwidth]{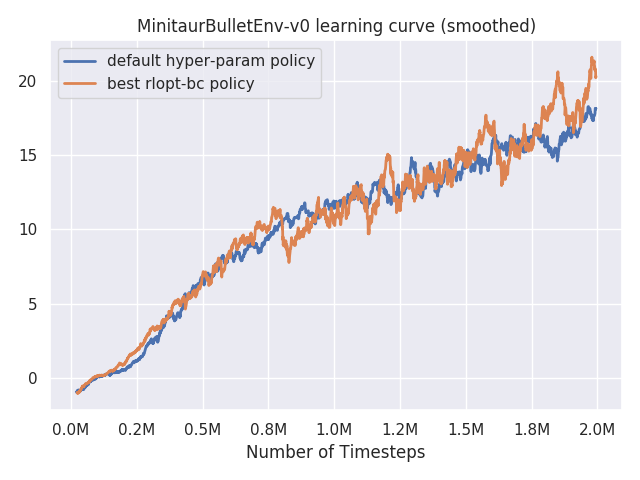}
		
	\end{tabular}
	
	\centering
	\includegraphics[height=0.395\textwidth]{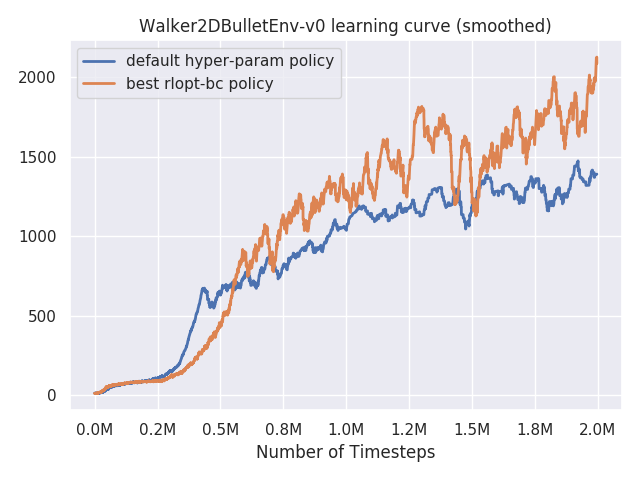}
	
	\caption{Different learning curves based on the best policies found after 10 meta-episodes using the proposed approach.}
	\label{fig:learning_curves}
\end{figure*}

Some of the learning curves for the best policies found with the proposed approach can be seen in Fig. \ref{fig:learning_curves} where, as a reference, they are compared with learning curves generated with hyper-parameters of RL Baselines Zoo repository \cite{rl-zoo} (a repository that has hyper-parameter configurations that produce converging policies in the environments validated in this subsection). A generated video showing examples of the agents trained with the generated policies at different meta-episodes (first meta-episode, best meta-episode and a meta-episode in-between the two) can be seen at \url{https://youtu.be/XDm2v6wt0-A}.

\begin{longtable}[ht!]{|l|l|c|}
	\caption{Computational times for the different optimizers in the five simulation environments considered.}
	\label{tbl:running_times_pybullet}\\
	\hline
	Environment  & Optimizer             & \multicolumn{1}{l|}{Avg. time (h)} \\ 
	\hline
	\endfirsthead
	
	Ant          & Random Search         & 7:52:06                        \\ \hline
	& RLOpt & 8:00:31                        \\ \hline
	& RLOpt-BC              & 8:33:41                        \\ \hline
	& TPE                   & 9:00:00*                       \\ \hline
	& SMAC             & 9:04:02                         \\ \hline
	Half-cheetah & Random Search         & 9:36:52                        \\ \hline
	& RLOpt & 9:40:00                        \\ \hline
	& RLOpt-BC              & 10:45:46                       \\ \hline
	& TPE                   & 11:00:00*                      \\ \hline
	& SMAC                & 10:43:40                       \\ \hline
	Hopper       & Random Search         & 4:24:43                        \\ \hline
	& RLOpt & 4:24:10                        \\ \hline
	& RLOpt-BC              & 4:59:01                        \\ \hline
	& TPE                   & 5:00:00*                       \\ \hline
	& SMAC                & 4:45:27                       \\ \hline

	Minitaur     & Random Search         & 32:21:04                       \\ \hline
	& RLOpt & 33:13:45                       \\ \hline
	& RLOpt-BC              & 34:12:55                       \\ \hline
	& TPE                   & 34:00:00*                      \\ \hline
	& SMAC                & 33:22:25                       \\ \hline
	Walker       & Random Search         & 7:52:12                        \\ \hline
	& RLOpt & 7:52:31                        \\ \hline
	& RLOpt-BC              & 10:06:07                       \\ \hline
	& TPE                   & 10:00:00*                      \\ \hline
	& SMAC                & 8:40:29                       \\ \hline
\end{longtable}

\subsection{Validation with broader search spaces}
\label{subsec_hyper_ranges}

As an additional study to test the sensitivity of the different optimizers to widespread ranges of hyper-parameters, two additional validations with increased ranges were performed.
These new experiments are similar to those in the previous Section, and focus in the Hopper Bullet environment, employing two additional sets of ranges (shown in Table \ref{tbl_chosen_hyperparam_ranges}) to make room for more exploration of the hyper-parameter space: the ``broader'' ranges, which considerably extends the search space of the previous Section, and the ``ample'' ranges, which extends the possible values for each hyper-parameter to near the maximum range.
For the ample set of ranges, this means that the algorithms can be instanced with hyper-parameters that are unlikely to be included in automatic optimization ranges, such as $\gamma$, that may take values of almost zero.
The only exception in this set is the value of the entropy coefficient, which was limited to take maximum values of $0.2$, as higher values will produce exploding gradients.

\begin{table}[b]
	\centering
	\caption{Different hyper-parameter minimum and maximum values for each set, used to test the sensibility of the predefined ranges.
			The ``original'' set is the set of ranges used in the preceding Section \ref{sec:pybullet_simulations}.	
	}
	\label{tbl_chosen_hyperparam_ranges}
	
	\begin{tabular}{|l|l|c|c|}
		\hline
		Hyper-parameter name        & Original ranges  & \multicolumn{1}{l|}{Broader ranges} & \multicolumn{1}{l|}{Ample ranges} \\ \hline
		Learning rate $\alpha_{lr}$ & $(1e-4, 1e-3)$   & $(1e-5, 1e-3)$                                & $(1e-7, 1e-3)$                              \\ \hline
		Discount factor $\gamma$    & $(0.8, 0.9999)$  & $(0.5, 0.9999)$                               & $(0.0001, 0.9999)$                          \\ \hline
		Clip range                  & $(0.1, 0.5)$     & $(0.01, 0.75)$                                & $(0.01, 0.99)$                              \\ \hline
		GAE $\lambda$               & $(0.85, 0.9999)$ & $(0.5, 0.9999)$                               & $(0, 0.9999)$                               \\ \hline
		Entropy coefficient         & $(0, 0.1)$       & $(0, 0.15)$                                   & $(0, 0.2)$                                  \\ \hline
		Value function coefficient  & $(0.5, 1)$       & $(0.25, 1)$                                   & $(0.1, 1)$                                  \\ \hline
	\end{tabular}
\end{table}

\begin{figure*}[h!]
	\centering
	\caption{Different maximum and cumulative rewards reached for each optimizer searching in the broad and ample hyper-parameter spaces, in the Hopper Bullet environment.}
	\label{fig_optimizers_increased_ranges}
	\begin{tabular}{ll}
		\includegraphics[height=0.39\linewidth]{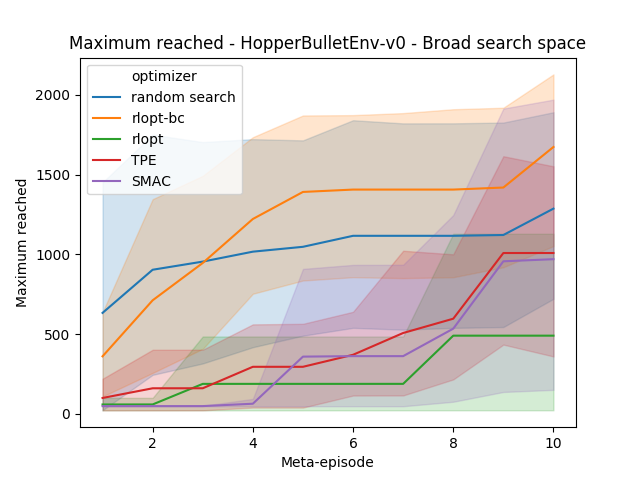}
		
		&
		
		\includegraphics[height=0.39\linewidth]{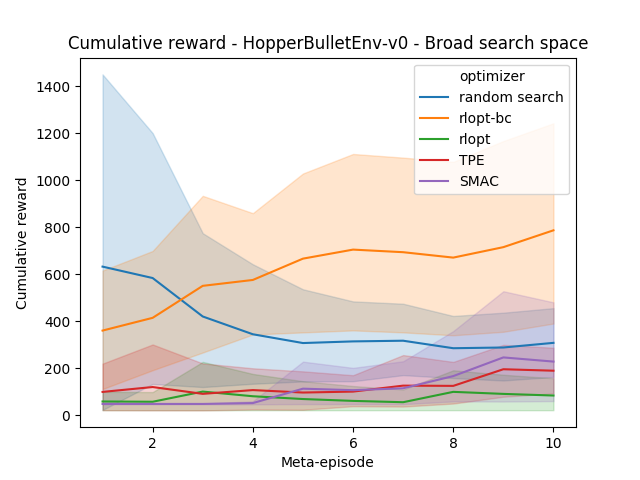}
		
		\\
		
		\includegraphics[height=0.39\linewidth]{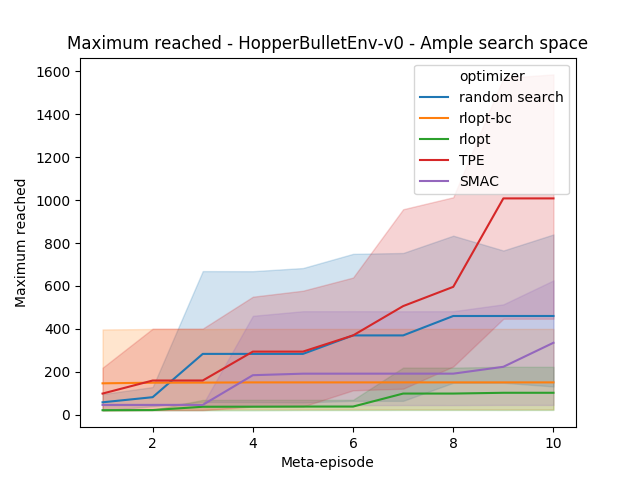}
		
		&
		
		\includegraphics[height=0.39\linewidth]{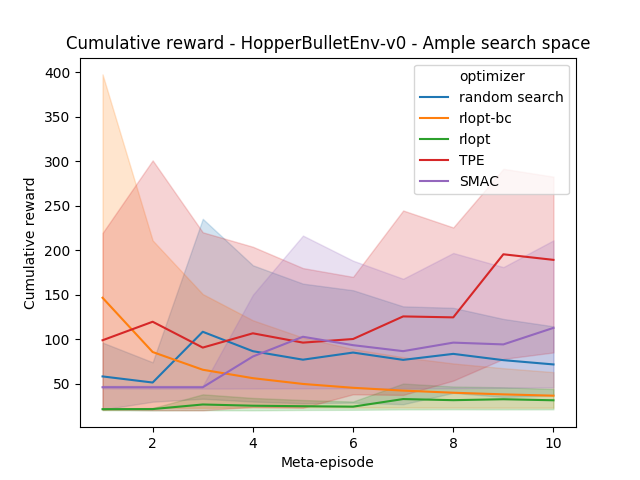}
		
	\end{tabular}
\end{figure*}

In the first experiment, the aim was to test all the optimizers of the previous Section \ref{sec:pybullet_simulations} to search for optimal hyper-parameters in the broad and ample search spaces.
Results of the simulations are depicted in Fig. \ref{fig_optimizers_increased_ranges}, where it is shown that, as it can be expected, an increase in the hyper-parameter range correlates with a decrease in the optimization performance. Also, increasing ranges significantly constitutes a major hindrance for policies to converge in the learning curve (this is best represented by the average cumulative rewards of each optimizer).
A noticeable aspect is that, compared with the results in the previous Section \ref{sec:pybullet_simulations}, there is also a higher separation in the correlation between the maximum reached at a certain meta-episode and their corresponding reward, which highlights that the different maxima found are heavily dependent on uncontrolled inputs, as its variance is very high (this is further increased when optimizing within ample search spaces).
Both range extensions give rise to a graceful degradation of all methods' performance, which is mainly due to the inherent sensitivity of hyper-parameter tuning in model-free RL, and amplified by the fact that both the policy and the value function are approximated using neural networks with evolving targets.
As it was done by limiting the range of the entropy coefficient, or selecting the range of the learning rate considering that the RL algorithm employs a neural network, it is a strongly recommended practice to include prior knowledge of the RL task when setting the hyper-parameter search,  by some form of transfer learning.

\begin{figure*}[h!]
	\centering
	\caption{Different maxima for the RLOpt-BC optimizer when adjusting one hyper-parameter at a time, for each of the three hyper-parameter ranges (shown in Table \ref{tbl_chosen_hyperparam_ranges}).}
	\label{fig_rlopt_only_one_hyp_maximum}
	\begin{tabular}{ll}
		\includegraphics[height=0.39\linewidth]{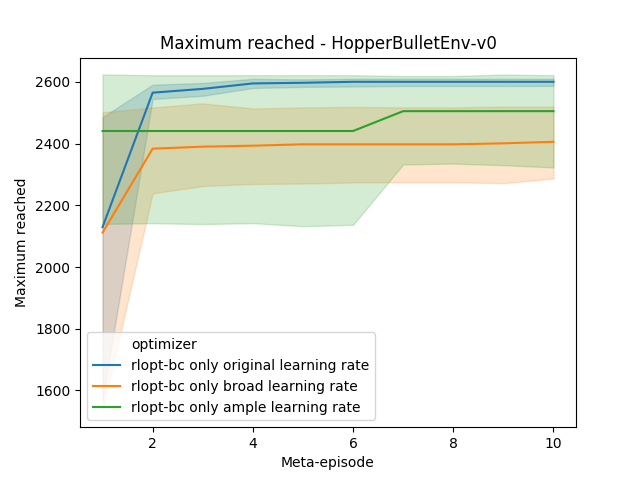}
		
		&
		
		\includegraphics[height=0.39\linewidth]{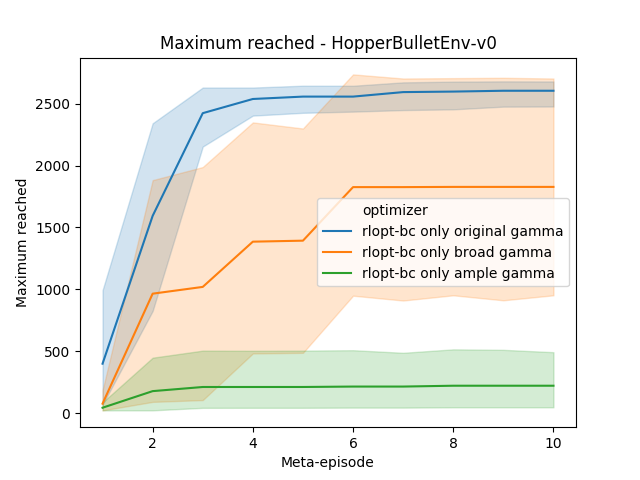}
		
		\\
		
		\includegraphics[height=0.39\linewidth]{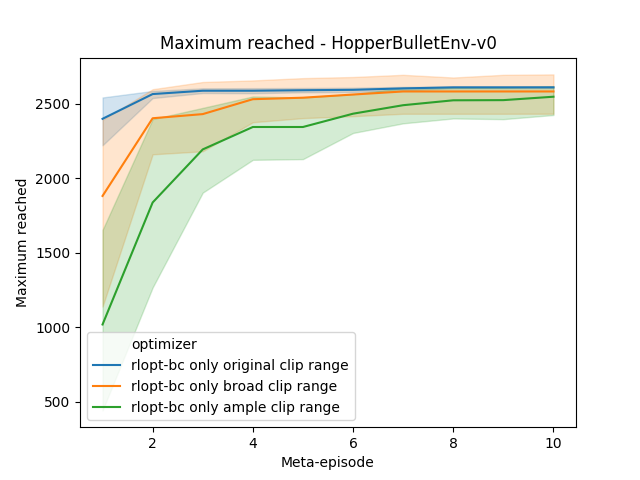}
		
		&
		
		\includegraphics[height=0.39\linewidth]{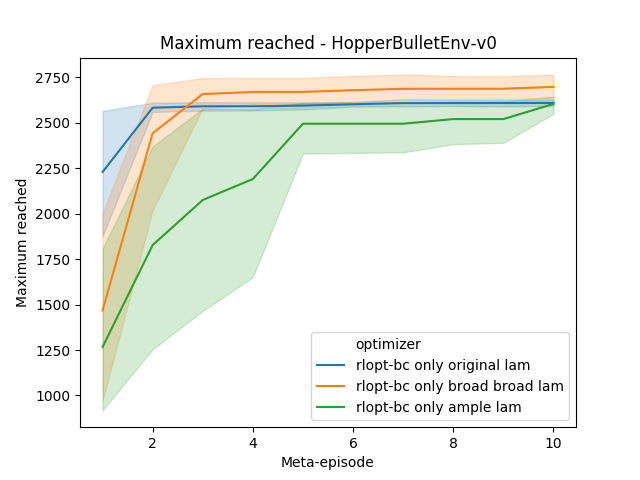}
		
		\\

		\includegraphics[height=0.39\linewidth]{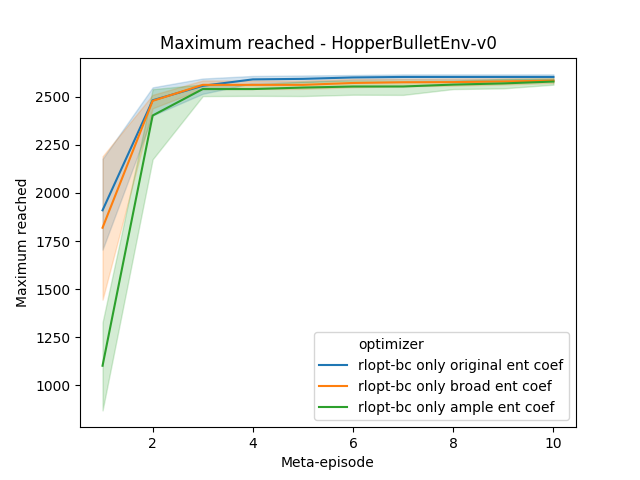}
		
		&
		
		\includegraphics[height=0.39\linewidth]{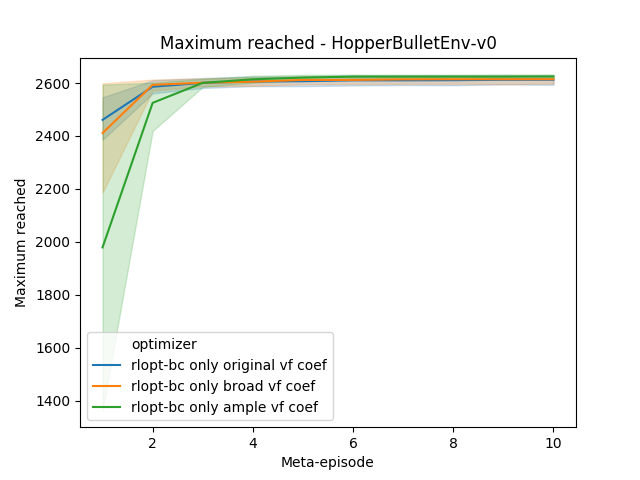}
		
	\end{tabular}
\end{figure*}

\begin{figure*}[h!]
	\centering
	\caption{Different cumulative rewards for the RLOpt-BC optimizer when adjusting one hyper-parameter at a time, for each of the three hyper-parameter ranges (shown in Table \ref{tbl_chosen_hyperparam_ranges}).}
	\label{fig_rlopt_only_one_hyp_cumulative}
	\begin{tabular}{ll}
		\includegraphics[height=0.39\linewidth]{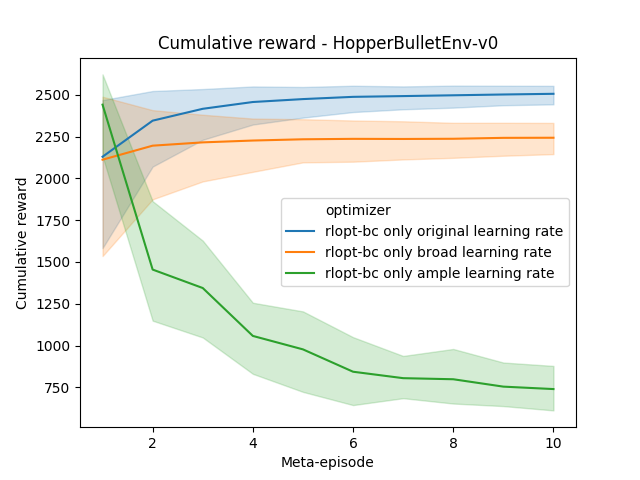}
		
		&
		
		\includegraphics[height=0.39\linewidth]{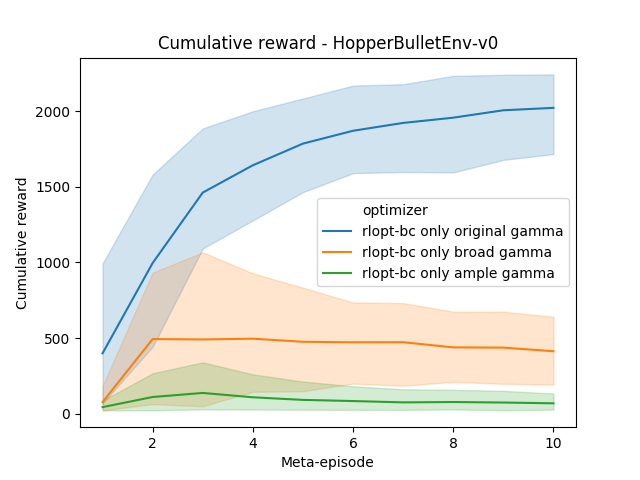}
		
		\\
		
		\includegraphics[height=0.39\linewidth]{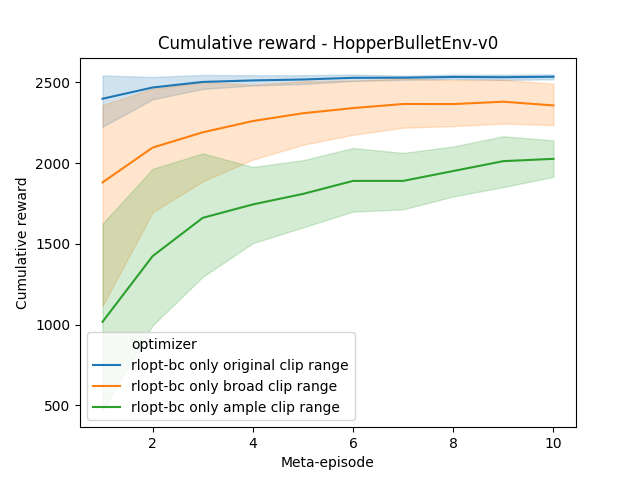}
		
		&
		
		\includegraphics[height=0.39\linewidth]{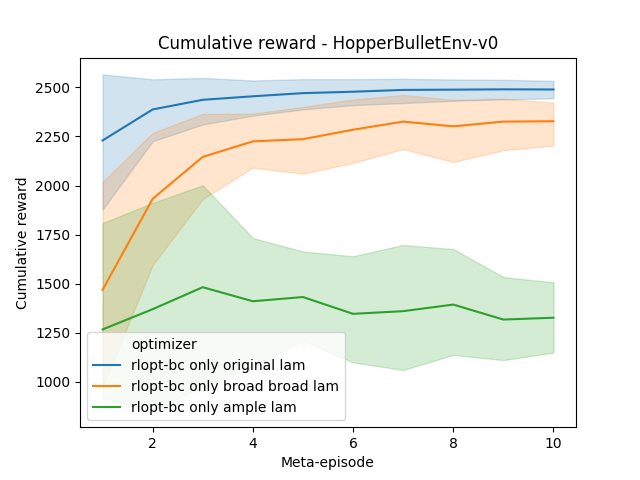}
		
		\\

		\includegraphics[height=0.39\linewidth]{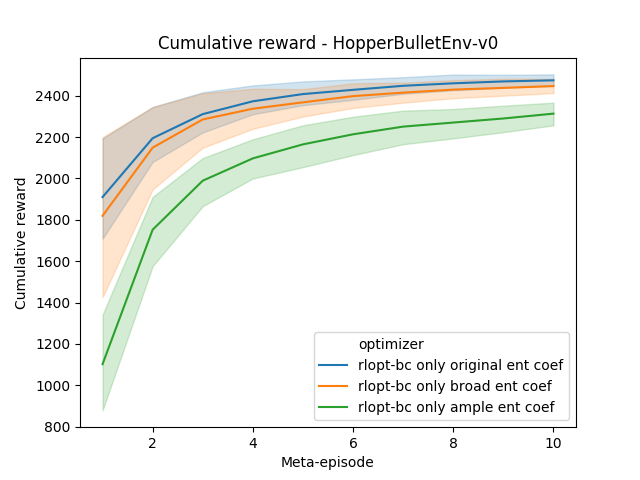}
		
		&
		
		\includegraphics[height=0.39\linewidth]{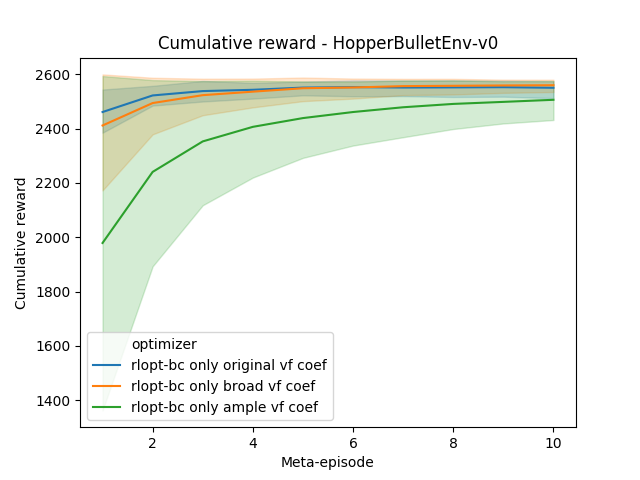}
		
	\end{tabular}
\end{figure*}

In the second experiment, an additional validation was done to test how the RLOpt-BC specifically performed when using different ranges, when tuning a single hyper-parameter at a time.
For this validation, three different experiments, (for the original, broad and ample sets), were performed for each one of the six hyper-parameters, which makes it possible to visualize to what extent the proposed optimizer adjusts one hyper-parameter while all the rest remain fixed.
The criteria to select the values of the fixed hyper-parameters was to set them at the same values that were used to generate the default hyper-parameter policies of Fig. \ref{fig:learning_curves} of the previous Section 3.1.
The results of these experiments are displayed in the curves of Fig. \ref{fig_rlopt_only_one_hyp_maximum} (maximum reached) and Fig. \ref{fig_rlopt_only_one_hyp_cumulative} (average cumulative rewards), where several aspects can be appreciated from them.
Firstly, the RLOpt-BC optimizer always converges towards good policies when adjusting one hyper-parameter at a time in the original ranges; therefore, tuning only one parameter does not decrease the overall performance. On the other hand, the sensitivity to extending the optimization ranges is more significant for some hyper-parameters compared to others which are far less sensitive; e.g., extending the search spaces of $\gamma$ and $\alpha_{lr}$ yields a higher impact in the performance than when extending the search space  for others.
Such an impact of expanding search ranges, as in the previous experiments, heavily depends on the range of the extension: for instance, adjusting the learning rates in the original and broad sets generate a correlation between the maxima found and the cumulative rewards obtained; this does not occur when the optimization ranges are extended toward the values of the ample set.

\subsection{Validation in BSuite}

An exhaustive validation was also performed using the Behavioral Suite (BSuite) \cite{osband_behaviour_2019} where, in order to validate the capabilities of the proposed algorithm in an array of vastly different environments, a comparison with two solid optimizers was made. A set of small environments (e.g., the classic Cartpole environment) was used to validate algorithms by evaluating their capabilities to solve different issues, such as their exploration or generalization capabilities. Such environments record agents' interaction and, when they finish, compute the normalized scores between 0 and 1 that denote agent performance.

As the optimal policy is known for many of the tasks due to their simplicity, the score is computed considering the regret regarding such policy. Additionally, each environment has 20 variants (random seeds), each presenting unique variations. These variations involve either small aspects, such as the initial state of the agent in the Cartpole or the arm that yields (on average) the optimal reward in a bandit agent, or more advanced aspects such as the noise that corrupts agent rewards (explained below) or the state space size of the environment. This helps to reduce the effect of stochasticity when evaluating agents, and to determine how agents perform in more difficult versions of the same environment. A complete list of the included environments and their description is included in the original paper \cite{osband_behaviour_2019}. The issues addressed in BSuite are the following:

\begin{itemize}
	\item \textbf{Basic learning}, consists of classic environments to test basic learning capabilities of an algorithm when solving tasks like N-armed bandit, Cartpole and MountainCar.
	
	\item \textbf{Noise}: basic learning experiments are repeated, this time testing how the agent learns when rewards are corrupted with a Gaussian noise $N(0, \sigma^2)$, where $\sigma$ takes different values from the list $[0.1, 0.3, 1, 3, 10]$, each one being present in 4 of the 20 available seeds.
	
	\item \textbf{Scale}: basic learning experiments are repeated in order to validate how the agent performs when rewards are multiplied by a fixed amount $c \in [0.1, 0.3, 1, 3, 10]$, being each value of $c$ present in 4 of the 20 seeds.
	
	\item \textbf{Exploration}: consists of environments that give sparse rewards. The reward function gives a considerably high reward to the final state, that can only be reached through a sequence of actions that give only small negative rewards. These environments test the agent exploration capabilities, where the most difficult environments cannot be solved without deep exploration mechanisms.
	
	\item \textbf{Credit assignment}: consists of three environments that test how the agent discovers the value of actions whose consequences are seen on a long-term basis, e.g., after a long chain of unrelated actions.
	
	\item \textbf{Memory}: consists of two simple environments that test how the agent can represent an internal state that can recall specific prior observations. The kind of agents that are capable to perform such representations usually have a recurrent unit such as a long short-term memory (LSTM), being a standard multi-layer perceptron network quite limited to form these internal states.
	
	\item \textbf{Generalization}: measures how the algorithm performs in an array of different environments.
\end{itemize}

BSuite's main focus is to compare the performance of different agents when solving the foregoing set of issues. It offers two main ways to summarize the results of different agents: a radar plot and a bar plot. The radar plot aggregates and displays the results of each learning agent in each of the 7 issues, where the result of each issue is the average of agents' score in the different environments that address said issue. The bar plot, on the other hand, shows how each agent performed individually in each environment, where each color represents the main issue for a given environment.

In this validation experiment, three hyper-parameter optimizers, RLOpt-BC, random search, and RLOpt were compared at finding hyper-parameters for each BSuite environment. For this experiment, Deep Q-Network (DQN) algorithm with Prioritized Experience Replay \cite{schaul_prioritized_2015-1} and Double Deep Q-Learning \cite{van_hasselt_deep_2015} extensions were used, and the optimized hyper-parameters with their respective ranges were the following:

\begin{itemize}
	\item Learning rate for the Adam optimizer $\alpha_{lr} \in (1e^{-5}, 1e^{-1})$.
	\item Discount factor $\gamma \in (0.8, 0.9999)$.
	\item Exploration rate $\varepsilon \in (0.1, 0.9)$, which determines the probability with which the agent selects a random action. In this DQN implementation, $\epsilon$ is linearly annealed from $\epsilon$ towards 0.
	\item $\alpha \in (0.4, 0.8)$, which determines the weight assigned to the highest priorities in the replay buffer.
	\item $\beta \in (0.4, 0.8)$, how importance sampling weights affects learning. In this DQN implementation, $\beta$ is linearly annealed from the selected value of $\beta$ towards 1, as importance sampling correction grows in importance at the end of learning \cite{schaul_prioritized_2015-1}.
\end{itemize}

To preserve consistency with how BSuite performance is evaluated in their environments, the value of $f(\vartheta)$ at each meta-episode was set as the BSuite score $\in [0,1]$ after each environment was run. In turn, each optimizer tuned hyper-parameters for each environment for 10 meta-episodes, and the episode results corresponding to the maximum of $f(\vartheta)$ for each agent was used to compute the radar and bar plots. Regarding RLOpt-BC, each rollout in the $\sigma_{EI-BC}(\vartheta)$ acquisition function (Algorithm \ref{alg:ei_rl}) was run for 2.5\% of the total episodes, and a size of 100 episodes (~0.1\% - 1\% of the total) was used for the demonstration set $\psi$.

\begin{figure*}
	\centering
	\includegraphics[width=0.7\textwidth]{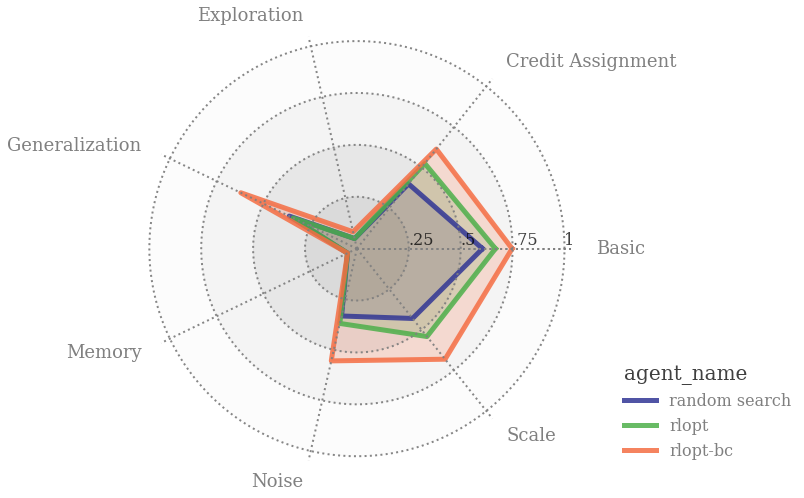}
	\caption{BSuite radar plot, showcasing how best policies found by each agent scored considering the different learning issues proposed by BSuite experiments.}
	\label{fig:bsuite_radar_plot}
\end{figure*}

\begin{figure*}
	\centering
	\includegraphics[width=\textwidth]{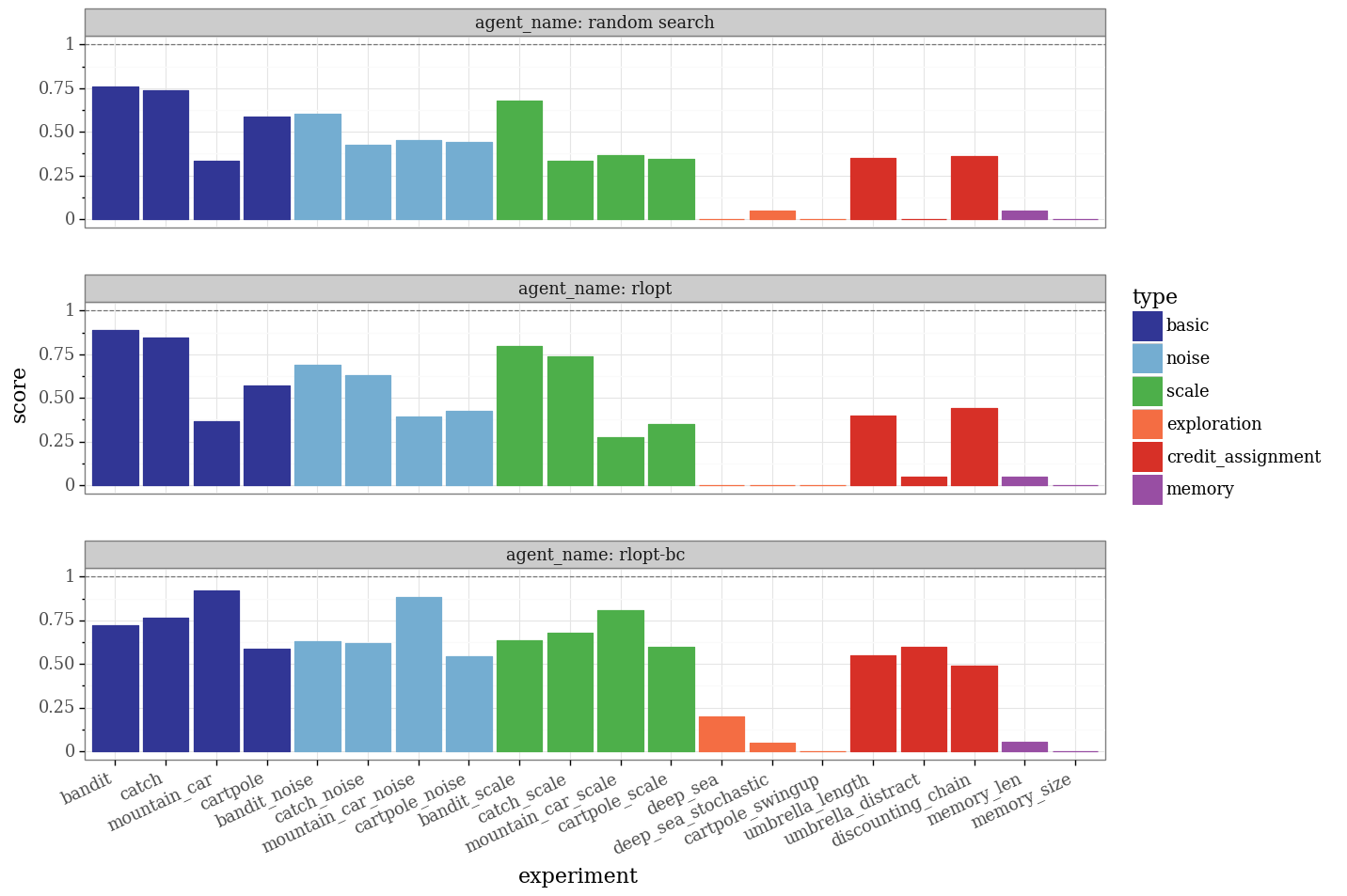}
	\caption{BSuite bar plot, showing how best policies found by each agent performed in each of the individual BSuite environments.}
	\label{fig:bsuite_bar_comparison}
\end{figure*}

Results of the BSuite experiments are displayed in Fig. \ref{fig:bsuite_radar_plot}, that summarizes an overview of the performance of the best trained policies at addressing the different issues, and Fig. \ref{fig:bsuite_bar_comparison}, that depicts the performance of the different policies in each individual environment.  As can be seen in the bar plot, the RLOpt-BC optimizer found policies that perform considerably better than the other two in problems such as MountainCar (in the basic, noise and scale setting), deep sea and umbrella distract. On the other hand, the three optimizers had similar performances in other basic, noise and scale environments, and they underperformed in deep exploration and memory environments such as Cartpole swing-up, where the DQN algorithm lacks deep exploration capabilities such as the proposed in Osband et al. (2019) \cite{JMLR:v20:18-339}, or recurrent units that enhances memory of recent actions, respectively. In the summary of the radar plot, RLOpt-BC is consistently better on average than the other two regarding the issues of noise, scale, generalization, credit assignment and basic, whereas it is slightly better at addressing exploration problems.

A closer look into environments that yielded significant performance differences among the three optimizers, or that can be of interest for gaining insight into the policies found the reader is referred to \ref{appdx_bsuite_analysis}.

As a side note, the MNIST set of problems (MNIST, MNIST noise and MNIST scale) offered in BSuite have been left out from the analysis. After extensive empirical validation that included both optimizing hyper-parameters and executing the same DQN algorithm in the same software libraries offered in the official BSuite repository, \url{https://github.com/deepmind/bsuite}, it was concluded that the MNIST environment for the latest BSuite version that was used in this work was not working correctly. A closer analysis of the reasons that motivated this decision are presented in \ref{mnist_analysis}.

\section{Concluding remarks} 
\label{sec:conclusion_and_future_works}

In this work, a novel approach for autonomous optimization of the hyper-parameters in a reinforcement learning agent has been presented. This approach involves taking into account the bias in data generation of hyper-parameter settings. To maximize the information content of actions taken and state transitions observed, a tight integration of Bayesian optimization with reinforcement learning using expected improvement acquisition function is proposed. As a result, the proposed approach leaves the user enough room to focus on the design of a meaningful reward function to guide the agent in its learning curve.

To speed up Bayesian optimization in the learning curve, different trajectories are stored and used as demonstrations to 1) increase the efficiency when selecting the next hyper-parameter vector $\vartheta$ using the current information gathered by the RL agent, and 2) perform behavioral cloning when evaluating a new configuration of hyper-parameters, in order to conveniently kick-start learning the optimal policy with a different bias. This gives rise to improved performance in fast converging towards optimal behavior with better control policies, compared with other optimization approaches.

Considering the results obtained, there are several aspects worth analyzing. Firstly, it was shown that RLOpt-BC performs consistently better than any of the alternative optimizers by resorting to bias based on hyper-parameter optimization while simultaneously converging to the optimal policy. Results are consistent for both on-policy and off-policy algorithms, both in a set of robotic simulation environments and in an array of small validation environments.

Regarding the simulations in PyBullet environments, the proposed approach performs consistently better in finding better sets of hyper-parameters than the other optimization approaches, and is suitable to find policies that are superior to the ones found with default hyper-parameters.
In terms of computing costs, it is fair to acknowledge that the proposed optimizer is a bit more demanding compared to random search or RLOpt, averaging a 13\% increase compared to random search. However, given that the proposed approach finds satisfactory solutions for $f(\vartheta)$ in the first 3-to 6 meta-episodes, the drawback of computational performance can be addressed by reducing the number of meta-episodes. This provides a reasonable trade-off between better hyper-parameter settings with more computational time, and also opens novel opportunities for improvement, for example by adding parallelization capabilities in the proposed acquisition function.
On the other hand, it was also shown that the proposed algorithm is perfectly capable of adjusting one hyper-parameter at a time, and that its performance is marginally sensitive to the size of the search space; however, this is mitigated by using prior knowledge of the task being optimized in the form of the hyper-parameter ranges, as it is a common practice when performing hyper-parameter optimization.

Regarding the performance of the obtained policies in the BSuite environments, it was shown that, on average, RLOpt-BC finds satisfactory policies for most environments, while having the same limitations of other competing algorithms in issues such as deep exploration or memory problems that require capabilities that the employed DQN algorithm does not have (for example, resorting to an LSTM unit).

As future research works, most of our current research efforts aim to validate the proposed approach in a full-scale industrial setting, where the optimizer must learn to autonomously set the hyper-parameters of an intelligent agent used for solving a real-time rescheduling task and manufacturing planning problems based on deep reinforcement learning. Also, the integration of the proposed approach with generative hypermodels of similar tasks is being addressed.


\section{Acknowledgments} 
\label{sec:acknowledgments}

This work was supported by the Universidad Tecnológica Nacional (UTN) and the National Scientific and Technical Research Council of Argentina (CONICET), and by the projects PAUTIFE0007699TC, EIUTIFE0004375TC and EIUTNVM0003581, all funded by the UTN. The authors want to thank Ezequiel Beccaría for the fruitful discussions and comments when defining the idea and experiments of this paper.


\section*{ORCID}

\begin{itemize}
	\item Juan Cruz Barsce: 0000-0003-4246-3727
	\item Jorge Andrés Palombarini: 0000-0002-9184-0632
	\item Ernesto Carlos Martínez:  0000-0002-2622-1579
\end{itemize}

\bibliographystyle{natbib}
\bibliography{Zotero.bib}

\begin{thebibliography}{10}

\bibitem{hutter_beyond_2015}
Hutter Frank, L{\"u}cke J{\"o}rg, {Schmidt-Thieme} Lars. Beyond {{Manual
  Tuning}} of {{Hyperparameters}}.  {\it KI - K\"unstliche Intelligenz.
  }2015;29(4):329--337.

\bibitem{bergstra_algorithms_2011}
Bergstra James~S., Bardenet R{\'e}mi, Bengio Yoshua, K{\'e}gl Bal{\'a}zs.
  Algorithms for {{Hyper}}-{{Parameter Optimization}}.  In:  {Shawe-Taylor} J.,
  Zemel R.~S., Bartlett P.~L., Pereira F., Weinberger K.~Q., eds. {\it Advances
  in {{Neural Information Processing Systems}} 24}, {Curran Associates, Inc.}
  2011 (pp. 2546--2554).

\bibitem{bergstra_random_2012}
Bergstra James, Bengio Yoshua. Random {{Search}} for {{Hyper}}-{{Parameter
  Optimization}}.  {\it Journal of Machine Learning Research.
  }2012;13(Feb):281--305.

\bibitem{snoek_practical_2012}
Snoek Jasper, Larochelle Hugo, Adams Ryan~P. Practical {{Bayesian
  Optimization}} of {{Machine Learning Algorithms}}.  In:  Pereira F., Burges
  C.~J.~C., Bottou L., Weinberger K.~Q., eds. {\it Advances in {{Neural
  Information Processing Systems}} 25}, {Curran Associates, Inc.} 2012 (pp.
  2951--2959).

\bibitem{zoph_neural_2016}
Zoph Barret, Le~Quoc~V.. Neural {{Architecture Search}} with {{Reinforcement
  Learning}}.  {\it arXiv:1611.01578 [cs]. }2016;.

\bibitem{loshchilov_cma-es_2016}
Loshchilov Ilya, Hutter Frank. {{CMA}}-{{ES}} for {{Hyperparameter
  Optimization}} of {{Deep Neural Networks}}.  {\it arXiv:1604.07269 [cs].
  }2016;.

\bibitem{sutton_reinforcement_2018}
Sutton Richard~S., Barto Andrew~G.. {\it Reinforcement {{Learning}}: {{An
  Introduction}}}.
\newblock Adaptive {{Computation}} and {{Machine Learning}}{Cambridge, Mass}:
  {MIT Press}; second~ed.2018.

\bibitem{mnih_human-level_2015}
Mnih Volodymyr, Kavukcuoglu Koray, Silver David, et al. Human-Level Control
  through Deep Reinforcement Learning.  {\it Nature. }2015;518(7540):529--533.

\bibitem{silver_mastering_2016}
Silver David, Huang Aja, Maddison Chris~J., et al. Mastering the Game of {{Go}}
  with Deep Neural Networks and Tree Search.  {\it Nature.
  }2016;529(7587):484--489.

\bibitem{silver_mastering_2017-1}
Silver David, Schrittwieser Julian, Simonyan Karen, et al. Mastering the Game
  of {{Go}} without Human Knowledge.  {\it Nature. }2017;550(7676):354.

\bibitem{vinyals_grandmaster_2019}
Vinyals Oriol, Babuschkin Igor, Czarnecki Wojciech~M., et al. Grandmaster Level
  in {{StarCraft II}} Using Multi-Agent Reinforcement Learning.  {\it Nature.
  }2019;:1--5.

\bibitem{lillicrap_continuous_2015}
Lillicrap Timothy~P., Hunt Jonathan~J., Pritzel Alexander, et al. Continuous
  Control with Deep Reinforcement Learning.  {\it arXiv:1509.02971 [cs, stat].
  }2015;.

\bibitem{henderson_deep_2017}
Henderson Peter, Islam Riashat, Bachman Philip, Pineau Joelle, Precup Doina,
  Meger David. Deep {{Reinforcement Learning}} That {{Matters}}.  {\it
  arXiv:1709.06560 [cs, stat]. }2017;.

\bibitem{islam_reproducibility_2017}
Islam Riashat, Henderson Peter, Gomrokchi Maziar, Precup Doina. Reproducibility
  of {{Benchmarked Deep Reinforcement Learning Tasks}} for {{Continuous
  Control}}.  {\it arXiv:1708.04133 [cs]. }2017;.

\bibitem{hester_deep_2017}
Hester Todd, Vecerik Matej, Pietquin Olivier, et al. Deep {{Q}}-Learning from
  {{Demonstrations}}.  {\it arXiv:1704.03732 [cs]. }2017;.

\bibitem{coumans2019}
Coumans Erwin, Bai Yunfei. {{PyBullet}}, a {{Python}} Module for Physics
  Simulation for Games, Robotics and Machine Learning  http://pybullet.org2016.

\bibitem{osband_behaviour_2019}
Osband Ian, Doron Yotam, Hessel Matteo, et al. Behaviour {{Suite}} for
  {{Reinforcement Learning}}.  In: ; 2019.

\bibitem{OpenAI_dota}
{OpenAI} . {{OpenAI Five}}  https://blog.openai.com/openai-five/2018.

\bibitem{shalev-shwartz_safe_2016}
{Shalev-Shwartz} Shai, Shammah Shaked, Shashua Amnon. Safe,
  {{Multi}}-{{Agent}}, {{Reinforcement Learning}} for {{Autonomous Driving}}.
  {\it arXiv:1610.03295 [cs, stat]. }2016;.

\bibitem{li_deep_2016}
Li~Jiwei, Monroe Will, Ritter Alan, Galley Michel, Gao Jianfeng, Jurafsky Dan.
  Deep {{Reinforcement Learning}} for {{Dialogue Generation}}.  {\it
  arXiv:1606.01541 [cs]. }2016;.

\bibitem{watkins_q-learning_1992}
Watkins Christopher J. C.~H., Dayan Peter. Q-Learning.  {\it Machine Learning.
  }1992;8(3-4):279--292.

\bibitem{schaul_prioritized_2015-1}
Schaul Tom, Quan John, Antonoglou Ioannis, Silver David. Prioritized
  {{Experience Replay}}.  {\it arXiv:1511.05952 [cs]. }2015;.

\bibitem{NIPS1999_1713}
Sutton Richard~S, McAllester David~A., Singh Satinder~P., Mansour Yishay.
  Policy Gradient Methods for Reinforcement Learning with Function
  Approximation.  In:  Solla S.~A., Leen T.~K., M{\"u}ller K., eds. {\it
  Advances in Neural Information Processing Systems 12}, {MIT Press} 2000 (pp.
  1057--1063).

\bibitem{schulman_high-dimensional_2015}
Schulman John, Moritz Philipp, Levine Sergey, Jordan Michael, Abbeel Pieter.
  High-{{Dimensional Continuous Control Using Generalized Advantage
  Estimation}}.  {\it arXiv:1506.02438 [cs]. }2015;.

\bibitem{schulman_proximal_2017}
Schulman John, Wolski Filip, Dhariwal Prafulla, Radford Alec, Klimov Oleg.
  Proximal {{Policy Optimization Algorithms}}.  {\it arXiv:1707.06347 [cs].
  }2017;.

\bibitem{argall_survey_2009}
Argall Brenna~D., Chernova Sonia, Veloso Manuela, Browning Brett. A Survey of
  Robot Learning from Demonstration.  {\it Robotics and Autonomous Systems.
  }2009;57(5):469--483.

\bibitem{ross_efficient_2010}
Ross Stephane, Bagnell Drew. Efficient {{Reductions}} for {{Imitation
  Learning}}.  In: :661--668; 2010.

\bibitem{rajeswaran_learning_2018}
Rajeswaran Aravind, Kumar Vikash, Gupta Abhishek, et al. Learning {{Complex
  Dexterous Manipulation}} with {{Deep Reinforcement Learning}} and
  {{Demonstrations}}.  In: ; 2018.

\bibitem{goecks_integrating_2020}
Goecks Vinicius~G, Gremillion Gregory~M, Lawhern Vernon~J, Valasek John,
  Waytowich Nicholas~R. Integrating {{Behavior Cloning}} and {{Reinforcement
  Learning}} for {{Improved Performance}} in {{Dense}} and {{Sparse Reward
  Environments}}.  In: :9; 2020; {New Zealand}.

\bibitem{mockus_application_1978}
Mo{\v c}kus Jonas, Tiesis Vytautas, Zilinskas A.. The Application of
  {{Bayesian}} Methods for Seeking the Extremum.  In:  Dixon L.C.W., Szego
  G.P., eds. {\it Towards {{Global Optimisation}} 2}, {North-Holand} 1978 (pp.
  117--129).

\bibitem{shahriari_taking_2016}
Shahriari B., Swersky K., Wang Ziyu, Adams R.P., {de Freitas} N.. Taking the
  {{Human Out}} of the {{Loop}}: {{A Review}} of {{Bayesian Optimization}}.
  {\it Proceedings of the IEEE. }2016;104(1):148--175.

\bibitem{rasmussen_gaussian_2008}
Rasmussen Carl~Edward, Williams Christopher K.~I.. {\it Gaussian Processes for
  Machine Learning}.
\newblock Adaptive Computation and Machine Learning{Cambridge, Mass.}: {MIT
  Press}; 3. print~ed.2008.

\bibitem{baptista_bayesian_2018}
Baptista Ricardo, Poloczek Matthias. Bayesian {{Optimization}} of
  {{Combinatorial Structures}}.  In: :462--471{PMLR}; 2018.

\bibitem{hutter_sequential_2011}
Hutter Frank, Hoos Holger~H., {Leyton-Brown} Kevin. Sequential
  {{Model}}-{{Based Optimization}} for {{General Algorithm Configuration}}.
  In: Lecture {{Notes}} in {{Computer Science}}:507--523{Springer, Berlin,
  Heidelberg}; 2011.

\bibitem{hansen_completely_2001}
Hansen Nikolaus, Ostermeier Andreas. Completely {{Derandomized
  Self}}-{{Adaptation}} in {{Evolution Strategies}}.  {\it Evolutionary
  Computation. }2001;9(2):159--195.

\bibitem{thornton_auto-weka_2013}
Thornton Chris, Hutter Frank, Hoos Holger~H., {Leyton-Brown} Kevin.
  Auto-{{WEKA}}: {{Combined Selection}} and {{Hyperparameter Optimization}} of
  {{Classification Algorithms}}.  In: {{KDD}} '13:847--855{ACM}; 2013; {New
  York, NY, USA}.

\bibitem{clark_moe:_2014}
Clark Scott, Liu Eric, Frazier Peter, Wang JiaLei, Oktay Deniz, Vesdapunt
  Norases. {{MOE}}: {{A}} Global, Black Box Optimization Engine for Real World
  Metric Optimization  https://github.com/Yelp/MOE2014.

\bibitem{bergstra_hyperopt:_2013}
Bergstra James, Yamins Dan, {David D. Cox} . Hyperopt: {{A Python Library}} for
  {{Optimizing}} the {{Hyperparameters}} of {{Machine Learning Algorithms}}.
  In:  {van der Walt} St{\'e}fan, Millman Jarrod, {Katy Huff} , eds. {\it
  Proceedings of the 12th {{Python}} in {{Science Conference}}}, :13--19; 2013.

\bibitem{li_hyperband_2017}
Li~Lisha, Jamieson Kevin, DeSalvo Giulia, Rostamizadeh Afshin, Talwalkar Ameet.
  Hyperband: {{A Novel Bandit}}-{{Based Approach}} to {{Hyperparameter
  Optimization}}.  {\it Journal of Machine Learning Research. }2017;18(1):52.

\bibitem{falkner_bohb_2018}
Falkner Stefan, Klein Aaron, Hutter Frank. {{BOHB}}: {{Robust}} and {{Efficient
  Hyperparameter Optimization}} at {{Scale}}.  In: :1437--1446{PMLR}; 2018.

\bibitem{akiba_optuna_2019}
Akiba Takuya, Sano Shotaro, Yanase Toshihiko, Ohta Takeru, Koyama Masanori.
  Optuna: {{A Next}}-Generation {{Hyperparameter Optimization Framework}}.  In:
  :2623--2631{ACM Press}; 2019; {Anchorage, AK, USA}.

\bibitem{white_greedy_2016}
White Martha, White Adam. A Greedy Approach to Adapting the Trace Parameter for
  Temporal Difference Learning.  In: {{AAMAS}} '16:557--565{International
  Foundation for Autonomous Agents and Multiagent Systems}; 2016; {Richland,
  SC}.

\bibitem{dabney_adaptive_2012}
Dabney William, Barto Andrew~G.. Adaptive Step-Size for Online Temporal
  Difference Learning.  In: {{AAAI}}'12:872--878{AAAI Press}; 2012; {Toronto,
  Ontario, Canada}.

\bibitem{barsce_towards_2017}
Barsce J.~C., Palombarini J.~A., Mart{\'i}nez E.~C.. Towards Autonomous
  Reinforcement Learning: {{Automatic}} Setting of Hyper-Parameters Using
  {{Bayesian}} Optimization.  In: :1--9; 2017.

\bibitem{chen_bayesian_2018}
Chen Yutian, Huang Aja, Wang Ziyu, et al. Bayesian {{Optimization}} in
  {{AlphaGo}}.  {\it arXiv:1812.06855 [cs, stat]. }2018;.

\bibitem{liessner_hyperparameter_2019}
Liessner Roman, Schmitt Jakob, Dietermann Ansgar, B{\"a}ker Bernard.
  Hyperparameter {{Optimization}} for {{Deep Reinforcement Learning}} in
  {{Vehicle Energy Management}}:.  In: :134--144{SCITEPRESS - Science and
  Technology Publications}; 2019; {Prague, Czech Republic}.

\bibitem{young_distributed_2020}
Young M.~Todd, Hinkle Jacob~D., Kannan Ramakrishnan, Ramanathan Arvind.
  Distributed {{Bayesian}} Optimization of Deep Reinforcement Learning
  Algorithms.  {\it Journal of Parallel and Distributed Computing.
  }2020;139:43--52.

\bibitem{francois-lavet_how_2016}
{Fran{\c c}ois-Lavet} Vincent, Fonteneau Raphael, Ernst Damien. How to
  {{Discount Deep Reinforcement Learning}}: {{Towards New Dynamic Strategies}}.
   {\it arXiv:1512.02011 [cs]. }2016;.

\bibitem{xu_meta-gradient_2018}
Xu~Zhongwen, {van Hasselt} Hado, Silver David. Meta-{{Gradient Reinforcement
  Learning}}.  {\it arXiv:1805.09801 [cs, stat]. }2018;.

\bibitem{zahavy_self-tuning_2020-1}
Zahavy Tom, Xu~Zhongwen, Veeriah Vivek, et al. A {{Self}}-{{Tuning
  Actor}}-{{Critic Algorithm}}.  {\it Advances in Neural Information Processing
  Systems. }2020;33:20913--20924.

\bibitem{NEURIPS2019_10ff0b5e}
Veeriah Vivek, Hessel Matteo, Xu~Zhongwen, et al. Discovery of Useful Questions
  as Auxiliary Tasks.  In:  Wallach H., Larochelle H., Beygelzimer A.,
  {dAlch{\'e}-Buc} F., Fox E., Garnett R., eds. {\it Advances in Neural
  Information Processing Systems}, {Curran Associates, Inc.}; 2019.

\bibitem{wang_beyond_2020}
Wang Yufei, Ye~Qiwei, Liu Tie-Yan. Beyond {{Exponentially Discounted Sum}}:
  {{Automatic Learning}} of {{Return Function}}.  {\it arXiv:1905.11591 [cs,
  stat]. }2020;.

\bibitem{NEURIPS2020_cceff8fa}
Zhou Wei, Li~Yiying, Yang Yongxin, Wang Huaimin, Hospedales Timothy. Online
  Meta-Critic Learning for off-Policy Actor-Critic Methods.  In:  Larochelle
  H., Ranzato M., Hadsell R., Balcan M.~F., Lin H., eds. {\it Advances in
  Neural Information Processing Systems}, :17662--17673{Curran Associates,
  Inc.}; 2020.

\bibitem{tang_online_2020}
Tang Yunhao, Choromanski Krzysztof. Online {{Hyper}}-Parameter {{Tuning}} in
  {{Off}}-Policy {{Learning}} via {{Evolutionary Strategies}}.  {\it
  arXiv:2006.07554 [cs, stat]. }2020;.

\bibitem{salimans_evolution_2017}
Salimans Tim, Ho~Jonathan, Chen Xi, Sidor Szymon, Sutskever Ilya. Evolution
  {{Strategies}} as a {{Scalable Alternative}} to {{Reinforcement Learning}}.
  {\it arXiv:1703.03864 [cs, stat]. }2017;.

\bibitem{jaderberg_population_2017}
Jaderberg Max, Dalibard Valentin, Osindero Simon, et al. Population {{Based
  Training}} of {{Neural Networks}}.  {\it arXiv:1711.09846 [cs]. }2017;.

\bibitem{mnih_asynchronous_2016}
Mnih Volodymyr, Badia Adri{\`a}~Puigdom{\`e}nech, Mirza Mehdi, et al.
  Asynchronous {{Methods}} for {{Deep Reinforcement Learning}}.  {\it
  arXiv:1602.01783 [cs]. }2016;.

\bibitem{franke_sample-efficient_2020}
Franke J{\"o}rg K.~H., Koehler Gregor, Biedenkapp Andr{\'e}, Hutter Frank.
  Sample-{{Efficient Automated Deep Reinforcement Learning}}.  In: ; 2020.

\bibitem{schmitt_off-policy_2019}
Schmitt Simon, Hessel Matteo, Simonyan Karen. Off-{{Policy Actor}}-{{Critic}}
  with {{Shared Experience Replay}}.  {\it arXiv:1909.11583 [cs, stat]. }2019;.

\bibitem{paine_hyperparameter_2020}
Paine Tom~Le, Paduraru Cosmin, Michi Andrea, et al. Hyperparameter
  {{Selection}} for {{Offline Reinforcement Learning}}.  {\it arXiv:2007.09055
  [cs, stat]. }2020;.

\bibitem{zhang_importance_2021}
Zhang Baohe, Rajan Raghu, Pineda Luis, et al. On the {{Importance}} of
  {{Hyperparameter Optimization}} for {{Model}}-Based {{Reinforcement
  Learning}}.  In: :4015--4023{PMLR}; 2021.

\bibitem{bouneffouf_online_2021}
Bouneffouf Djallel, Claeys Emmanuelle. Online {{Hyper}}-{{Parameter Tuning}}
  for the {{Contextual Bandit}}.  In: :3445--3449; 2021.

\bibitem{stable-baselines}
Hill Ashley, Raffin Antonin, Ernestus Maximilian, et al. Stable Baselines
  https://github.com/hill-a/stable-baselines2018.

\bibitem{tan_sim--real_2018}
Tan Jie, Zhang Tingnan, Coumans Erwin, et al. Sim-to-{{Real}}: {{Learning Agile
  Locomotion For Quadruped Robots}}.  {\it arXiv:1804.10332 [cs]. }2018;.

\bibitem{jamieson_non-stochastic_2016}
Jamieson Kevin, Talwalkar Ameet. Non-Stochastic {{Best Arm Identification}} and
  {{Hyperparameter Optimization}}.  In: :240--248{PMLR}; 2016.

\bibitem{smac-2017}
Lindauer Marius, Eggensperger Katharina, Feurer Matthias, Falkner Stefan,
  Biedenkapp Andr{\'e}, Hutter Frank. {\it {{SMAC}} v3: {{Algorithm}}
  Configuration in Python. } 2017.

\bibitem{kingma_adam_2017}
Kingma Diederik~P., Ba~Jimmy. Adam: {{A Method}} for {{Stochastic
  Optimization}}.  {\it arXiv:1412.6980 [cs]. }2017;.

\bibitem{rl-zoo}
Raffin Antonin. {{RL}} Baselines Zoo
  https://github.com/araffin/rl-baselines-zoo2018.

\bibitem{van_hasselt_deep_2015}
{van Hasselt} Hado, Guez Arthur, Silver David. Deep {{Reinforcement Learning}}
  with {{Double Q}}-Learning.  {\it arXiv:1509.06461 [cs]. }2015;.

\bibitem{JMLR:v20:18-339}
Osband Ian, Roy Benjamin~Van, Russo Daniel~J., Wen Zheng. Deep Exploration via
  Randomized Value Functions.  {\it Journal of Machine Learning Research.
  }2019;20(124):1--62.

\bibitem{li_hyperband_2016-1}
Li~Lisha, Jamieson Kevin, DeSalvo Giulia, Rostamizadeh Afshin, Talwalkar Ameet.
  Hyperband: {{A Novel Bandit}}-{{Based Approach}} to {{Hyperparameter
  Optimization}}.  {\it arXiv:1603.06560 [cs, stat]. }2016;.

\bibitem{lecun_gradient-based_1998}
Lecun Y., Bottou L., Bengio Y., Haffner P.. Gradient-Based Learning Applied to
  Document Recognition.  {\it Proceedings of the IEEE. }1998;86(11):2278--2324.

\end{thebibliography}

\newpage

\appendix

\section{Experiments with additional optimizers}
\label{appdx_additional_experiments}

In this section of the appendix, the experiments that were run with other optimizers and were not included in the plots of Section \mbox{\ref{sec:pybullet_simulations}}, are shown.

In particular, three other optimization alternatives were tried:

\begin{itemize}
	
	\item Random sampler from Optuna \mbox{\cite{akiba_optuna_2019}}, which samples points randomly, similar to random search, incorporating pruning mechanisms for unpromising trials.

	\item Covariance Matrix Adaptation Evolution Strategy (CMA-ES) \mbox{\cite{loshchilov_cma-es_2016}}, which was also implemented with the Optuna backend.

	\item Bayesian Optimization with Hyperband (BOHB), which combines Bayesian optimization with Hyperband \mbox{\cite{li_hyperband_2016-1}}, and it was also implemented in the Optuna backend.
	
\end{itemize}

\begin{figure*}[hb!]
	\centering
	\begin{tabular}{ll}
		\includegraphics[height=0.39\textwidth]{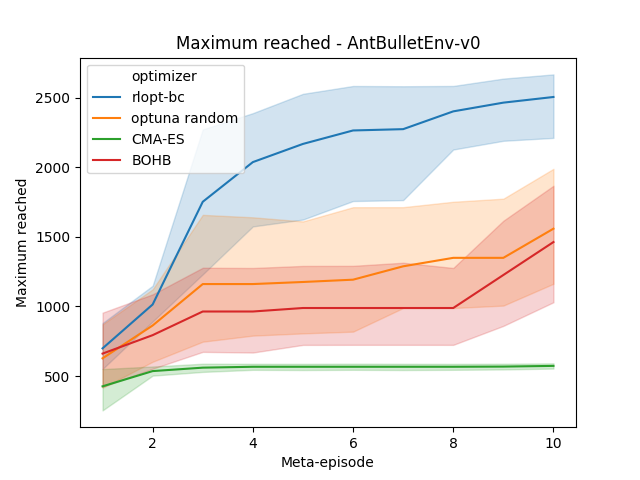}
		&
		\includegraphics[height=0.39\textwidth]{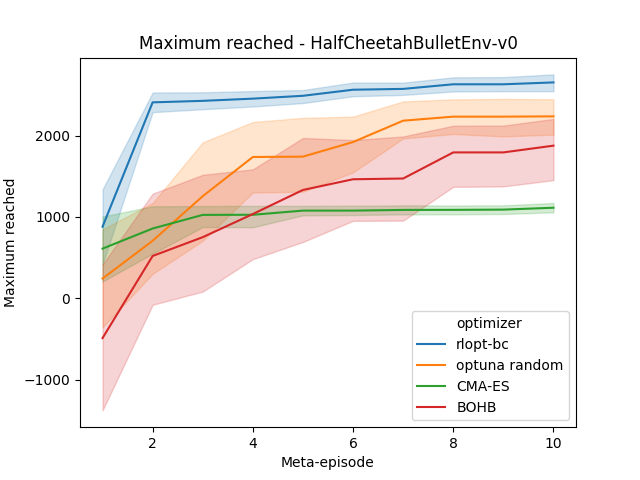} \\
		
		\includegraphics[height=0.39\textwidth]{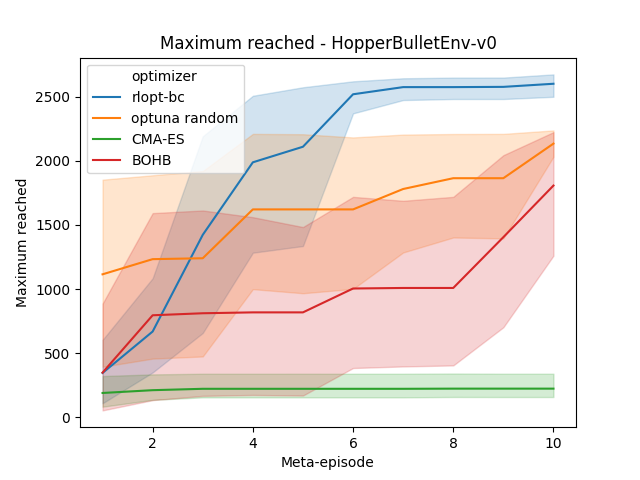}
		&
		\includegraphics[height=0.39\textwidth]{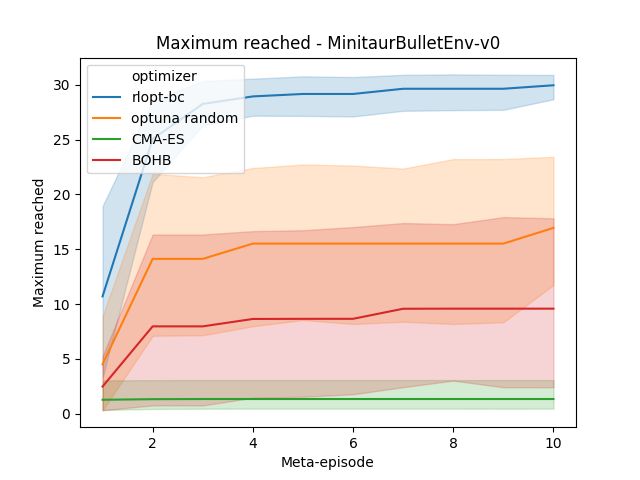}
		
	\end{tabular}
	
	\centering
	\includegraphics[height=0.39\textwidth]{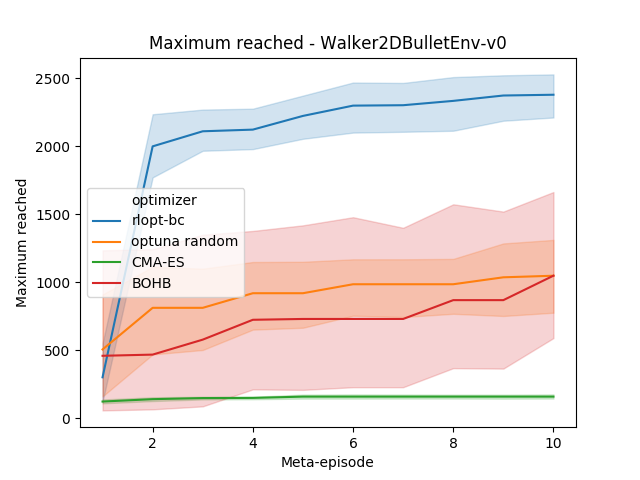}
	
	\caption{Comparison of the average maximum reached at each meta-episode for each of the optimizers, considering six different executions.}
	\label{fig:maximums_reached_otheropt}
\end{figure*}

\begin{figure*}[hb!]
	\centering
	\begin{tabular}{ll}
		\includegraphics[height=0.39\textwidth]{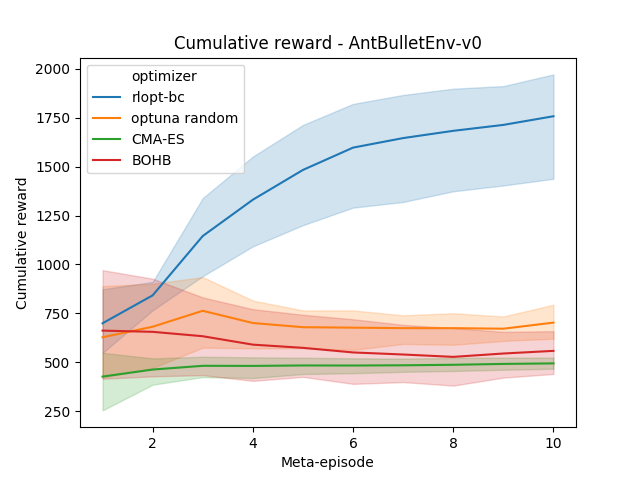}
		&
		\includegraphics[height=0.39\textwidth]{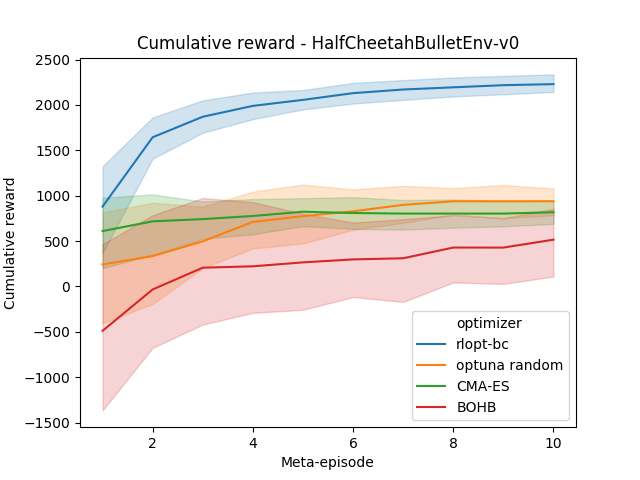} \\
		
		\includegraphics[height=0.39\textwidth]{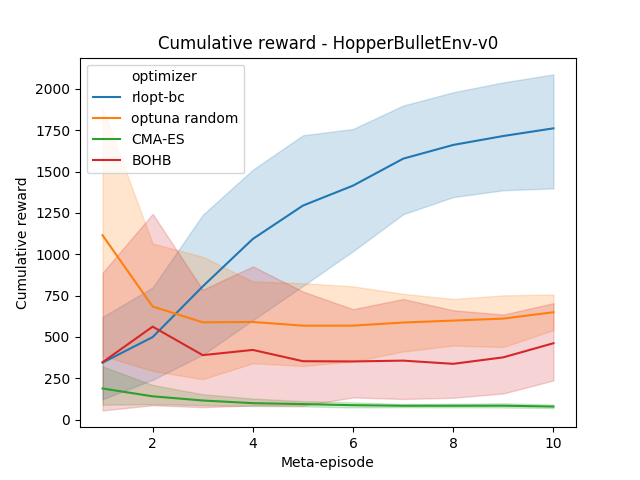}
		&
		\includegraphics[height=0.39\textwidth]{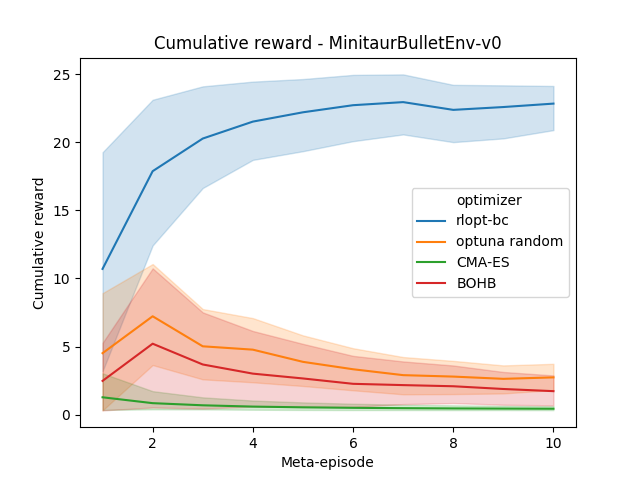}
		
	\end{tabular}
	
	\centering
	\includegraphics[height=0.39\textwidth]{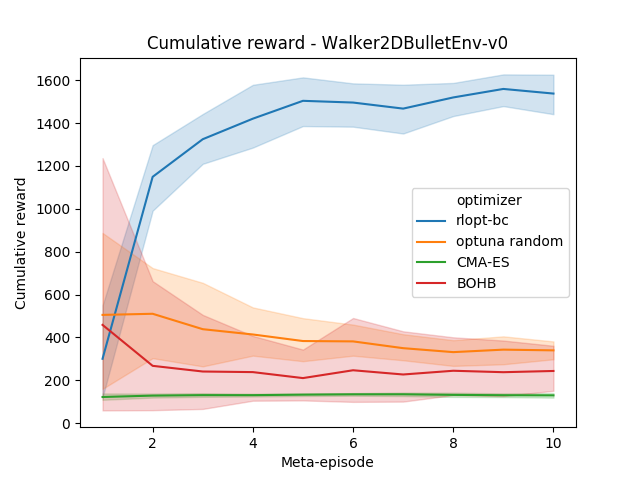}
	
	\caption{Comparison of the average cumulative rewards reached at each meta-episode for each of the optimizers, considering six different executions.}
	\label{fig:cumulative_rewards_otheropt}
\end{figure*}

The results of the application of the three optimizers, plus RLOpt-BC (as reference) in the five PyBullet environments is shown in Fig. \mbox{\ref{fig:maximums_reached_otheropt}} (in terms of their maximum reached) and Fig. \mbox{\ref{fig:cumulative_rewards_otheropt}} (in terms of their cumulative rewards).
It can be appreciated that among the three optimizers, the best was the random sampler from Optuna, which performed similar to the random search optimizer.
The second best optimizer was the combination of Bayesian optimization with Hyperband, that did not surpass the other Bayesian optimization counterparts, RLOpt and TPE.
Lastly, the optimizer that yielded the worst results was CMA-ES, producing poor policies in all the environments.
This may appear surprising, however, this is rather expected with population-based methods, as they typically require large number of samples in order to have good performances.

\begin{figure*}[h!]
	\centering
	\caption{Different maximum and cumulative rewards reached for each optimizer searching in the broad and ample hyper-parameter spaces, in the Hopper Bullet environment.}
	\label{fig_optimizers_increased_ranges_otheropt}
	\begin{tabular}{ll}
		\includegraphics[height=0.39\linewidth]{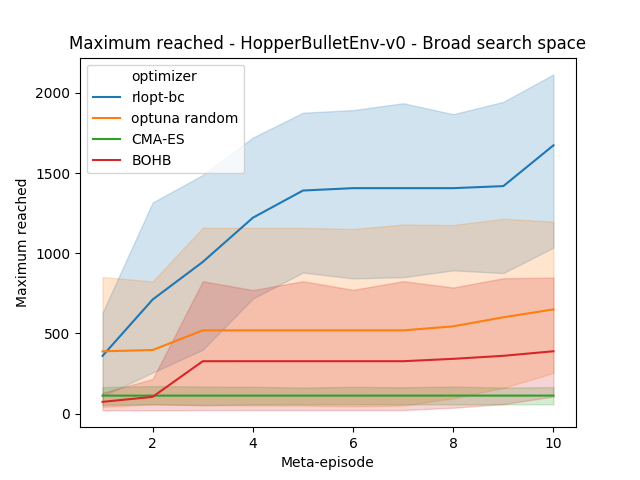}
		
		&
		
		\includegraphics[height=0.39\linewidth]{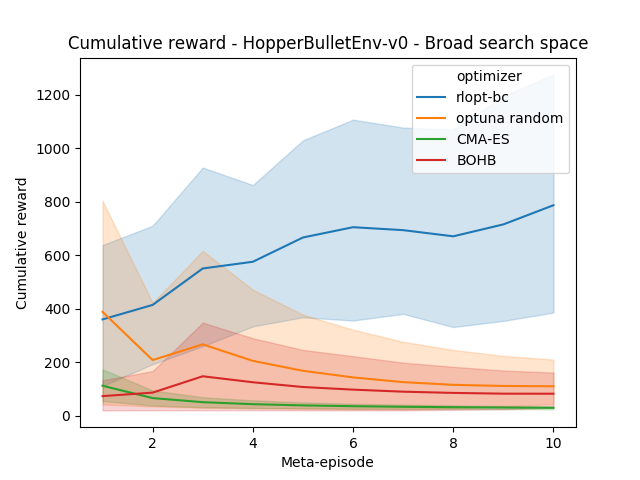}
		
		\\
		
		\includegraphics[height=0.39\linewidth]{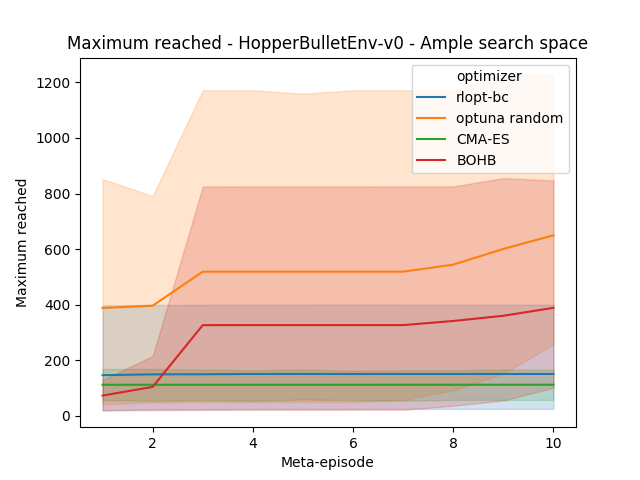}
		
		&
		
		\includegraphics[height=0.39\linewidth]{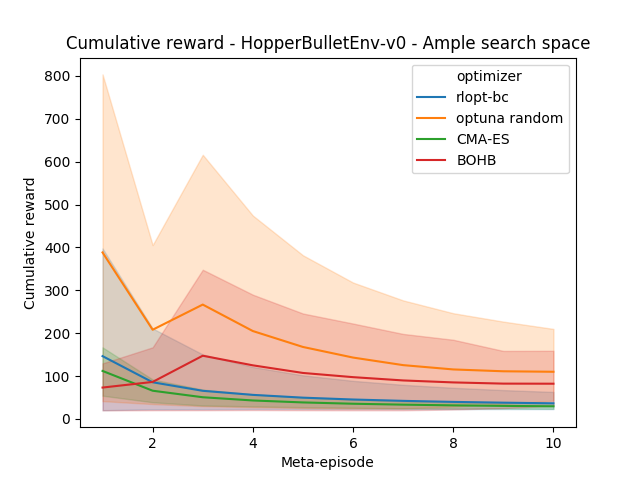}
		
	\end{tabular}
\end{figure*}

\section{Closer look at several BSuite experiments}
\label{appdx_bsuite_analysis}

In this section of the appendix, the BSuite experiments that yielded higher differences or are of special interest are analyzed in greater detail.

\begin{figure*}
	\centering
	\includegraphics[width=\textwidth]{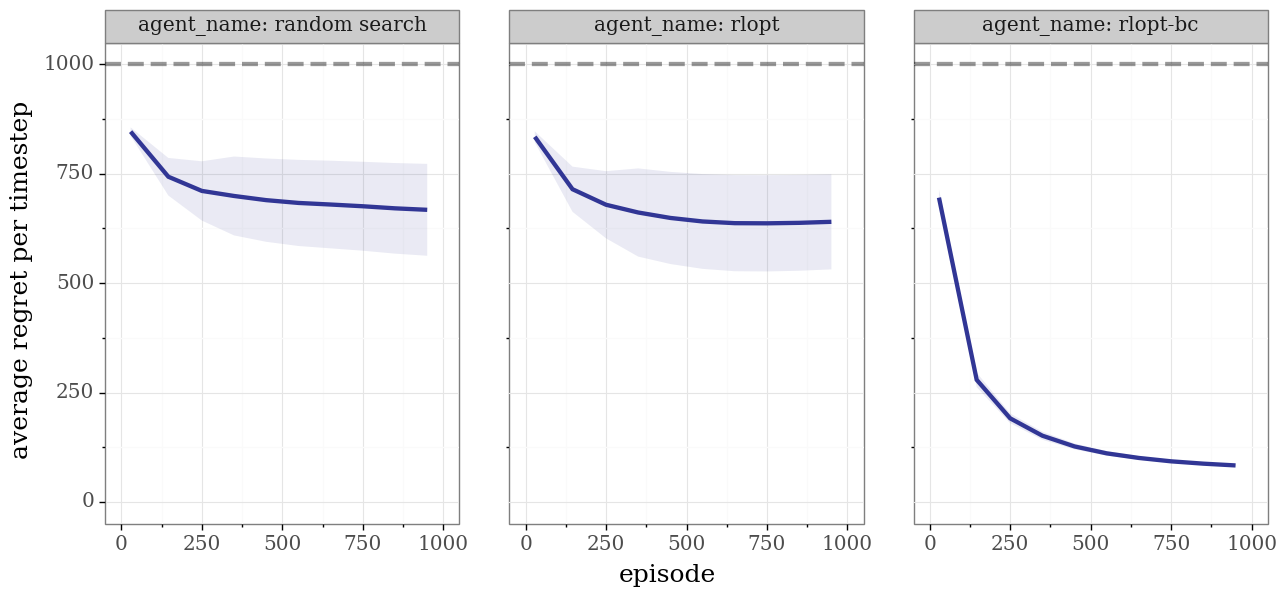}
	\caption{Average regret of 20 random seeds for the random search, Bayesian optimization, and RLOpt-BC optimizers in the MountainCar environment.}
	\label{fig:mountain_car_analysis}
\end{figure*}

\begin{itemize}
	\item MountainCar: classic problem that consists of an underpowered car that lies in the bottom of a valley, and the driver agent must learn to reach the top of the right hill. At each time-step, the agent receives a reward signal of -1. The average performance for each of the 20 seeds in terms of regret (respect to the optimal policy, where lower is better) for each optimizer can be seen in Fig. \ref{fig:mountain_car_analysis}, where the upper dashed line represents the average expected regret of a random agent. This is the environment where the RLOpt-BC optimizer reached the highest difference compared to random search and RLOpt.
\end{itemize}

\begin{figure*}
	\centering
	\includegraphics[width=\textwidth]{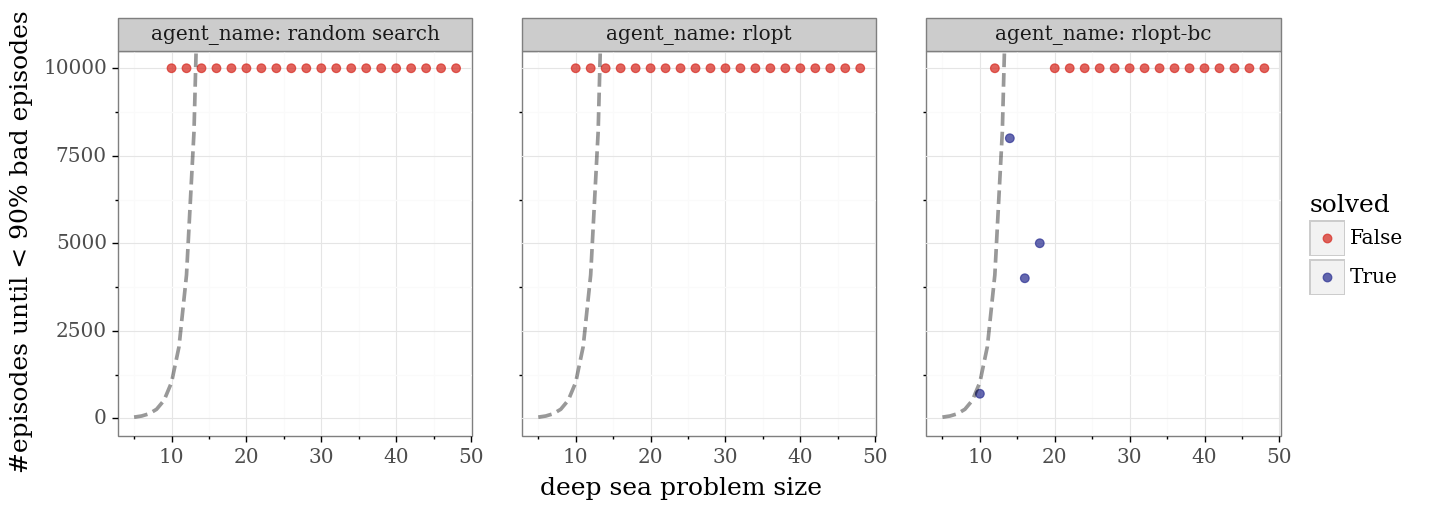}
	\caption{Number of episodes required for the agent to reach a regret which is less than 90\% for the best policies found by the three optimizers considered in the deep sea environment. The task is considered solved if the agent manages to avoid such a regret threshold in less than 10000 episodes (and thus it is marked with a blue dot); otherwise, it is marked by a red dot.}
	\label{fig:deep_sea_analysis}
\end{figure*}

\begin{itemize}
	\item Deep sea: this is a learning problem with special focus on the exploration capabilities, where an agent is situated in an $NxN$ grid with the start location in the top-left of the grid and it can move down-left and down-right. A down-left move has no cost while moving down-right has associated a small negative reward of $-0.01/N$, except for the bottom-right state that yields a reward of +1. An episode finishes when the agent reaches one of the bottom states. This is a small, yet difficult problem where the agent must learn to explore the grid traversing a seemingly suboptimal path until it reaches the positive reward at the bottom-right, to eventually learn the optimal policy of always moving down-right.
	
	Fig. \ref{fig:deep_sea_analysis} depicts the number of episodes needed when applying the best policies found by each optimizer to reach an average regret $< 0.9$ (which is a metric where the deep sea problem is considered as solved). The performance is compared to the dashed curve of $2^N$ episodes, a baseline of the expected scaling for deep exploration agents. Only RLOpt-BC is able to find policies that solve the problem (denoted by blue dots) for DQN in 4 of the first 5 sizes greater than 10.  Similarly to the other two optimizers, RLOpt-BC is unable to find policies for grids of greater sizes, as DQN lacks mechanisms for deep exploration.
	
	There is a variant of this problem, called \textit{stochastic deep sea}, where transitions to the right are performed with probability $(1 - 1/N)$, and the reward of the bottom states is corrupted by Gaussian noise $N(0,1)$. This is a very difficult problem to solve without deep exploration mechanisms, and none of the optimizers managed to find policies for any grid size other than the first.
\end{itemize}

\begin{figure*}
	\centering
	\includegraphics[width=\textwidth]{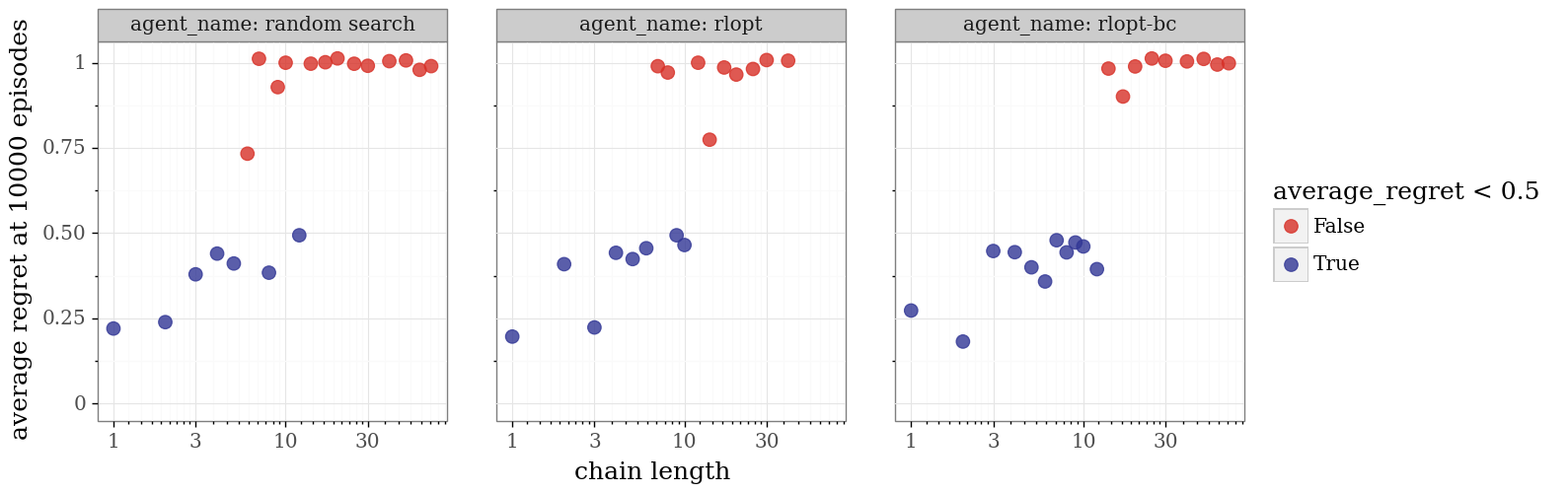}
	\caption{Average regret (lower values are better) for 10000 episodes, according to the length of the chain of intermediate states, in the umbrella length environment. Blue dots indicate that the task is considered solved; otherwise, it is indicated by a red dot.}
	\label{fig:umbrella_length_analysis}
\end{figure*}

\begin{figure*}
	\centering
	\includegraphics[width=\textwidth]{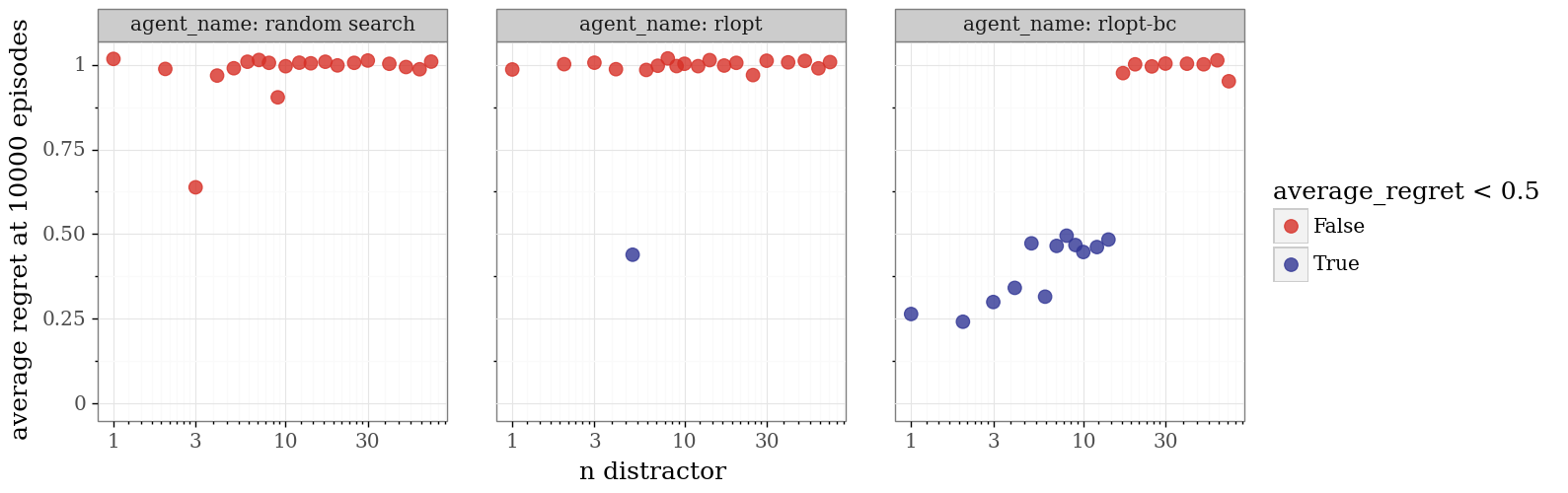}
	\caption{Average regret (lower values are better) for 10000 episodes in the umbrella distract environment, according to the number of distractor features. A blue dot indicates that the task is considered solved; otherwise, a red dot is used.}
	\label{fig:umbrella_distract_analysis}
\end{figure*}

\begin{itemize}
	\item Umbrella environments: \textit{umbrella} problems address the task of correctly crediting the actions and state features that were responsible for the reward received when reaching the final state. The state is composed of a fixed set of features that states if the umbrella is needed, if the umbrella has been taken, the number of remaining states to visit, and other unrelated random features. When each episode starts, the agent observes a state representing the forecast, and decides in the first action if it is taking an umbrella or not. Then there is a chain of unrelated states where the agent receives random rewards $\in \{-1,1\}$, and in the final state it receives +1 if it made the right decision by taking (or not) the umbrella; otherwise, it receives -1. In the first variant, called \textit{umbrella length}, each random seed corresponds to a different size of the chain of unrelated states. In Fig. \ref{fig:umbrella_length_analysis}, the average regret for each size of the chain can be observed for each of the optimizers considered, where a regret of $< 0.5$ indicates that the problem is solved (which is denoted by a blue point).
	
	There is another variant of the umbrella problem called \textit{umbrella distract}, where the amount of unrelated states is kept fixed, but the amount of unrelated features increases with each random seed. The average regret obtained by the best policies obtained by the different optimizers in this variant is shown in Fig. \ref{fig:umbrella_distract_analysis}
\end{itemize}

\section{Closer look of MNIST experiments}
\label{mnist_analysis}

In this Appendix, a closer look into the details that motivated the decision to not consider MNIST in the main analysis is presented. MNIST \cite{lecun_gradient-based_1998} is a classic machine learning data set that consists of classifying 60000 digital images of handwritten digits from 0 to 9. MNIST is typically solved with simple feed-forward neural network architectures (e.g., by flattening each $28 \times 28$ image into a 784 vector), without the need of convolutional layers. In Osband et al. (2019) \cite{osband_behaviour_2019}, its usage was extended to enable solving it by an RL algorithm, casting it as a bandit problem that returns +1 reward if the algorithm took the action that corresponded to the correct digit, and -1 otherwise.

When random search, RLOpt and RLOpt-BC optimizers attempted to find policies, it was noticeable that, on average, each of the best policies found scored 20\% accuracy in the basic, noise and scale MNIST environments, which is slightly above the performance of a random agent. Considering the simplicity of the problem, several experiments were carried out to find out if the reasons for such surprising under-performance were due to the optimization algorithms, the DQN implementation used, or the environment. The validation experiments included using a standalone PPO and DQN algorithm, and recreating the same virtual environment that was used in the experiments following the installation steps \url{https://github.com/deepmind/bsuite#installation} and running in Tensorflow and JAX baseline DQN implementations as in Osband et al. (2019) \cite{osband_behaviour_2019} (at the time of writing, the latest BSuite version is 0.3.5). As no better performance could be obtained even using the code of the BSuite repository, it was concluded that the reasons for this sub-par performance were external to the proposed algorithm, and thus such experiment was left out from the main analysis. The Fig. \ref{fig:mnist_dispersion} below shows the different MNIST scores obtained with the three optimizers, the performance of the standalone DQN algorithm using the code from the BSuite repository (light blue bar), and the score that was reported in Osband et al. (2019) \cite{osband_behaviour_2019} (red bar). 

\begin{figure*}
	\centering
	\includegraphics[width=0.8\textwidth]{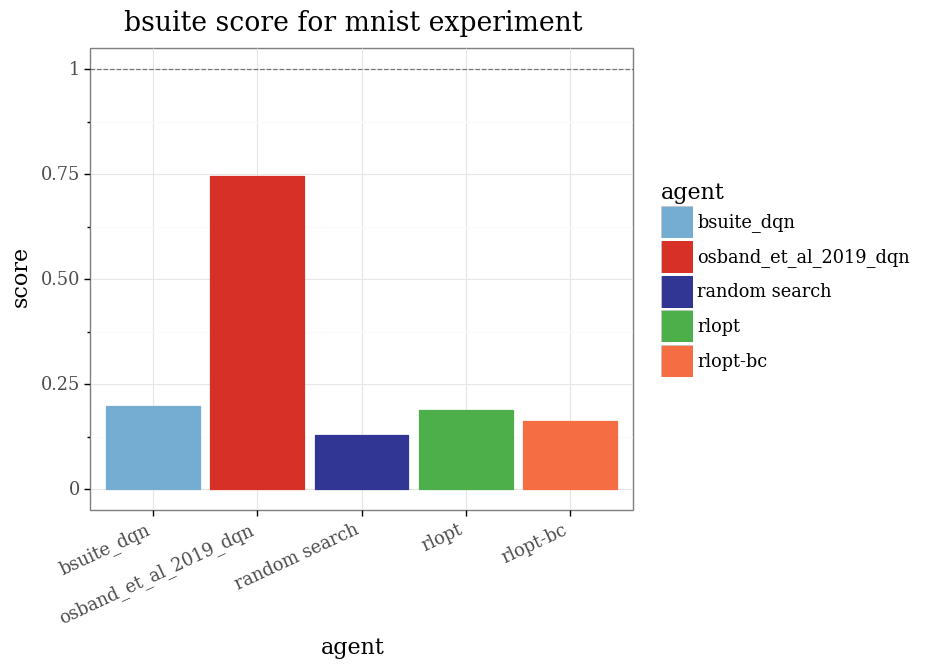}
	\caption{Different scores on MNIST bandit environment reached by random search, RLOpt and RLOpt-BC optimizers, along with the score reached with the DQN agent.}
	\label{fig:mnist_dispersion}
\end{figure*}

\end{document}